\let\mypdfximage\pdfximage
\def\pdfximage{\immediate\mypdfximage}

\documentclass[journal,onecolumn,10pt]{IEEEtran1}

\makeatletter
\def\markboth#1#2{\def\leftmark{\@IEEEcompsoconly{\sffamily}\MakeUppercase{\protect#1}}%
\def\rightmark{\@IEEEcompsoconly{\sffamily}\MakeUppercase{\protect#2}}}
\makeatother

\let\labelindent\relax

\usepackage[T1]{fontenc}
\usepackage{tgtermes}
\usepackage{lscape}
\usepackage{pdfpages}
\usepackage{marvosym}
\usepackage{longtable}
\usepackage{graphics}
\usepackage{graphicx}
\usepackage{caption}
\usepackage[english]{babel}
\usepackage{subfig}
\usepackage{epsfig}
\usepackage{color}
\usepackage{multirow}
\usepackage{mathrsfs}
\usepackage{textcomp}
\usepackage{amsfonts}
\usepackage{amsmath}
\usepackage{amssymb}
\usepackage{nccmath}
\usepackage[numbers,square,sort&compress,comma]{natbib}
\usepackage{chapterbib}
\usepackage[ruled,vlined]{algorithm2e}
\usepackage{setspace}
\usepackage{verbatim}
\usepackage{wrapfig}
\usepackage{dblfloatfix}
\usepackage{comment}
\usepackage[strict]{changepage}
\usepackage{ragged2e}
\usepackage{microtype}
\usepackage{enumitem}
\usepackage{nccmath}
\usepackage{hyperref}
\usepackage{doi}
\usepackage{afterpage}
\usepackage{wrapfig}
\usepackage[nameinlink]{cleveref}

\DeclareCaptionFont{singlespacing}{\setstretch{1}}
\numberwithin{figure}{section}
\renewcommand{\thefigure}{\arabic{section}.\arabic{figure}}

\hypersetup{bookmarks=false,bookmarksopen=false,pdfpagemode=empty,pdfstartview=}

\renewcommand \thesection{\arabic{section}}
\numberwithin{equation}{section}

\title{\singlespacing\sf\huge Annotating Motion Primitives for Simplifying\\ Action Search in Reinforcement Learning}
\author{Isaac J. Sledge, \emph{Member, IEEE}, Darshan W. Bryner, \emph{Member, IEEE},\\ and Jos\'{e} C. Pr\'{i}ncipe, \emph{Life Fellow, IEEE}%
\thanks{\fontdimen2\font=1.6pt Isaac J. Sledge is the Senior Machine Learning Scientist and Delores M. Etter Assistant Secretary of the Navy Emergent Engineer with the Advanced Signal Processing and Automated Target Recognition Branch, Naval Surface Warfare Center, Panama City, FL, USA (email: isaac.j.sledge@navy.mil).  He the director of the Machine Intelligence Defense (MIND) lab at the Naval Sea Systems Command.}%
\thanks{\fontdimen2\font=1.6pt Darshan W. Bryner the Branch Head of the Advanced Signal Processing and Automated Target Recognition Branch, Naval Surface Warfare Center, Panama City, FL 32407 (email: darshan.bryner@navy.mil).}%
\thanks{\fontdimen2\font=1.6pt Jos\'{e} C. Pr\'{i}ncipe is the Don D. and Ruth S. Eckis Chair and Distinguished Professor with the Department of Electrical and Computer Engineering and the Department of Biomedical Engineering, University of Florida, Gainesville, FL 32611, USA (email: principe@cnel.ufl.edu).  He is the director of the Computational NeuroEngineering Laboratory (CNEL) at the University of Florida.\vspace{0.1cm}}%
\thanks{The work of the first and last authors was funded by grant N00014-19-WX-00636 (Marc Steinberg) from the US Office of Naval Research (ONR).  The first author was additionally supported by in-house laboratory independent research (ILIR) grant N00014-19-WX-00687 (Frank Crosby) from the US ONR and two Naval Innovation in Science and Engineering (NISE) grants from the US Naval Sea Systems Command.}%
\thanks{\copyright 2021 IEEE.  Personal use of this material is permitted.  Permission from IEEE must be obtained for all other uses, in any current or future media, including reprinting/republishing this material for advertising or promotional purposes, creating new collective works, for resale or redistribution to servers or lists, or reuse of any copyrighted component of this work in other works.}
}
\begin{document}

\maketitle
\RaggedRight\parindent=1.5em
\fontdimen2\font=2.2pt
\vspace{-1.55cm}\begin{abstract}\normalsize\singlespacing
\vspace{-0.25cm}{\small{\sf{\textbf{Abstract}}}}---Reinforcement learning in large-scale environments is challenging due to the many possible actions that can be taken in specific situations.  We have previously developed a means of constraining, and hence speeding up, the search process through the use of motion primitives; motion primitives are sequences of pre-specified actions taken across a state series.  As a byproduct of this work, we have found that if the motion primitives' motions and actions are labeled, then the search can be sped up further.  Since motion primitives may initially lack such details, we propose a theoretically viewpoint-insensitive and speed-insensitive means of automatically annotating the underlying motions and actions.  We do this through a differential-geometric, spatio-temporal kinematics descriptor, which analyzes how the poses of entities in two motion sequences change over time.  We use this descriptor in conjunction with a weighted-nearest-neighbor classifier to label the primitives using a limited set of training examples.

In our experiments, we achieve high motion and action annotation rates for human-action-derived primitives with as few as one training sample.  We also demonstrate that reinforcement learning using accurately labeled trajectories leads to high-performing policies more quickly than standard reinforcement learning techniques.  This is partly because motion primitives encode prior domain knowledge and preempt the need to re-discover that knowledge during training.  It is also because agents can leverage the labels to systematically ignore action classes that do not facilitate task objectives, thereby reducing the action space.  

\end{abstract}%
\begin{IEEEkeywords}\normalsize\singlespacing
\vspace{-1.25cm}{{\small{\sf{\textbf{Index Terms}}}}---Motion annotation, motion classification, constraint learning, motion constraint, reinforcement learning}
\end{IEEEkeywords}
\IEEEpeerreviewmaketitle
\allowdisplaybreaks
\singlespacing

\vspace{-0.6cm}\subsection*{\small{\sf{\textbf{1$\;\;\;$Introduction}}}}\addtocounter{section}{1}

Optimally making decisions in the presence of uncertainty is crucial for intelligent agents.  Reinforcement learning \cite{SuttonRS-book1998a} addresses this problem by directing agents to maximize an expected return provided by the environment.  That is, context-specific action choices are considered, in a trial-and-error-based fashion, to iteratively construct an action-selection policy with good long-term performance.

Reinforcement learning has proved effective for handling a wide range of application domains.  However, it can be challenging to learn good policies for certain applications.  This is often due to the sheer number of actions that can be chosen for different contexts.  While advances have been made in transfer learning \cite{KonidarisG-conf2006a,LazaricA-conf2008a,LazaricA-conf2010a,HigginsI-conf2017a,BarretoA-coll2017a,TehYW-coll2017a}, in interpolating experiences across contexts \cite{BoyanJA-coll1995a,TesauroG-coll1996a,TsitsiklisJN-coll1996a,SuttonRS-coll1999a,MahadevanS-coll2006a,BhatnagarS-coll2009a}, and in other areas, it still can take a great amount of time to adequately evaluate possible action choices and thus arrive at a near-cost-optimal policy. 

Our aim is to constrain the action-search process in reinforcement learning to improve policy performance.  We propose to do this through macro-action representations called motion primitives.  Motion primitives are fragments of movement sequences that contain one or more variable-duration actions \cite{DudekD-conf2003a,JakelR-conf2010a,CohenBJ-conf2010a,CohenBJ-conf2011a,PowellMJ-conf2012a}.  By defining feasible movement trajectories, speed-ups in the action-search process can realized.  This is partly because the primitives encode task-specific kinematics knowledge.  They hence relieve the agent of needing to generate similar motions through trial and error.  As well, using motion primitives reduces the complexity of the action-search process.  The search process becomes a problem of discerning what macro-action works well across a series of states versus what many, potentially unique, actions work well \cite{ThrunSB-coll1995a}. 

A pair of problems needs to be solved to exploit the capabilities of motion trajectories in reinforcement learning.  For good-performing policies to be uncovered, a significantly diverse set of motion primitives must be available.  The first problem hence involves how to extract motion primitives to create a sufficiently large database.  The second problem involves how to employ the primitives.  

As part of the latter problem, we have found it is effective if the reinforcement-learning process has some high-level knowledge of what takes place during an action sequence.  An example would be the type of action that occurs during the primitive and the associated limb motions.  With such information, motion sequences that would be poorly suited for completing certain tasks can be quickly filtered.  This allows the agent to focus solely on primitives that are determined to be task relevant.  The action-search process will hence be locally constrained.  This should reduce the time needed to learn a good policy compared to the case where the agent can choose any non-labeled primitive.  

Our contribution in this paper is an differential-geometric scheme for labeling both the motions and the actions that occur in motion trajectories so that they can be more effectively employed for hierarchical reinforcement learning \cite{BartoAG-coll2003}.  We focus on the offline labeling of primitives derived from human actions \cite{PoppeR-jour2010a,WeinlandD-jour2011a}.  The scheme that we propose can also label primitives defined by arbitrary curves and surfaces.  

Our annotation scheme makes use of temporal pose self-similarities \cite{HancockE-book2013} to quantify how related an unlabeled primitive is to a labeled one.  More specifically, we consider a kinematics-based similarity measure that quantifies the similarity between the pose and location of an entity at one instant in time with its pose and location at other instants.  As a part of its construction, the metric can determine spatial and temporal correspondences for poses in different sequences, which is needed to produce good similarity values \cite{KaickO-jour2011a}.  We, however, simplify aspects of the correspondence problem to lessen the computational burden of using the measure. 

We compute exhaustive pairwise similarities using this measure.  We then compare similarities from an unlabeled sequence to those from a set of labeled motion examples.  In doing so, it becomes possible to transfer labels \cite{BrynerD-jour2014a,HasanbelliuE-jour2014a} using simple classifiers, like nearest neighbors.  We demonstrate that our annotation scheme performs well for several benchmark datasets.  This holds even when a single class-specific example is provided.  

The differential-geometric nature of our similarity measure comes with several benefits that aid in pre-processing motion primitives for reinforcement learning purposes.  It naturally encodes the notion of pose deformation, which feature-based approaches cannot always mimic well.  It can also be made near-viewpoint invariant, is not overly influenced by the speed at which an action is performed, and can be made somewhat insensitive to noise in the actor pose estimation.  As we show, these properties make our approach well suited to annotating primitives with only a few training examples.  State-of-the-art deep-learning-based schemes require many times more training samples to achieve the same performance, in contrast.  We can additionally define an exponential mapping to linearize the kinematics configuration space, permitting the extraction of statistics \cite{BrynerD-jour2017a}.  This permits defining the average representation of an action class, for instance. 

\begin{figure*}[!t]
   \hspace{0.175cm}\includegraphics[width=6.1in]{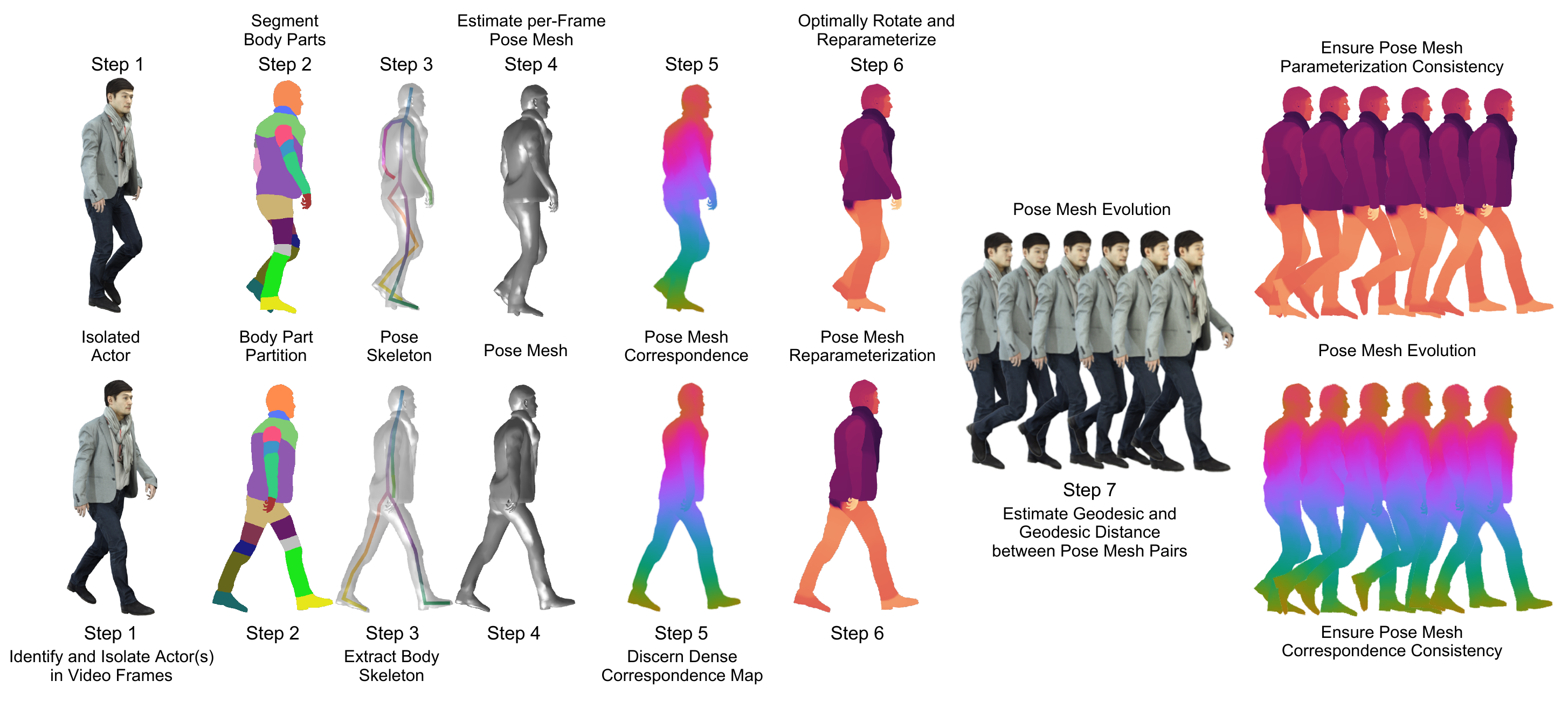}\vspace{-0.15cm}
   \caption{\fontdimen2\font=1.55pt\selectfont An overview of the pose similarity quantification process for three-dimensional shapes.  For a given video sequence, one or more actors are identified using a {\sc TrackRCNN} network (step 1).  An image-based body-part segmentation map is then formed via an {\sc FCN-SCP} network (step 2).  A body skeleton is also simultaneously estimated using a {\sc UNet} architecture (step 3).  Both the segmentation map and the body skeleton are fed into a {\sc DensePose} network to obtain a discrete, three-dimensional pose mesh per video frame (step 4).  For a pair of poses, a dense, whole-mesh correspondence map is obtained (step 5).  The correspondence map is used to rotate the pose meshes, thereby simplifying the geodesic path inference.  It is also used to reparameterize the pose meshes so that the meshes have an equivalent representation (step 6).  This facilitates comparing shapes.  Without a consistent parameterization, equivalent shapes defined by different meshes would have non-zero dissimilarity.  This would artificially bias the shape similarities and hence complicate primitive annotation.  For a pair of pose meshes with a consistent parameterization, a geodesic path is found on the pose orbifold (step 7).  The corresponding geodesic distance, described by our metric, specifies the amount of deformation from one pose mesh to another.  It hence quantifies pose similarity.  This geodesic is specified by solving a finite-dimensional boundary-value problem using a {\sc DGM} network.  While the metric is reparameterization invariant, it is often possible to ensure parameterization consistency across the geodesic, which improves the solution recovery rate.  Each step is repeated for every pair of frames in a video sequence.\vspace{-0.4cm}}
  \label{fig:fig0}
\end{figure*}

This paper is organized as follows.  We first give a brief overview of the relevant motion primitive literature in section 2.  After this, in section 3, we discuss the objectives of our primitive labeling methodology, which include detecting changes in underlying motions and characterizing the motions over time.  Both objectives are crucial for creating a robust annotation approach.  We show that they can be addressed by considering pose kinematics similarities.  The remainder of this section focuses on how we quantify pose similarity.  This process is summarized in \cref{fig:fig0}.  Associated theoretical and practical properties are covered in the online appendix. 

In section 4, we qualitatively demonstrate that the returned similarity values can differentiate between actions.  We show that small to moderate changes in the actions do not alter the similarities greatly.  We additionally show that differences in viewpoints and actor body types have almost no impact on the returned similarity values.  The similarities also vary greatly across action types.  Such findings suggest that our descriptor may be suitable for a wide range of actions and individuals.

\begin{figure}[t!]
   \hspace{0.95cm}\includegraphics[width=5.525in]{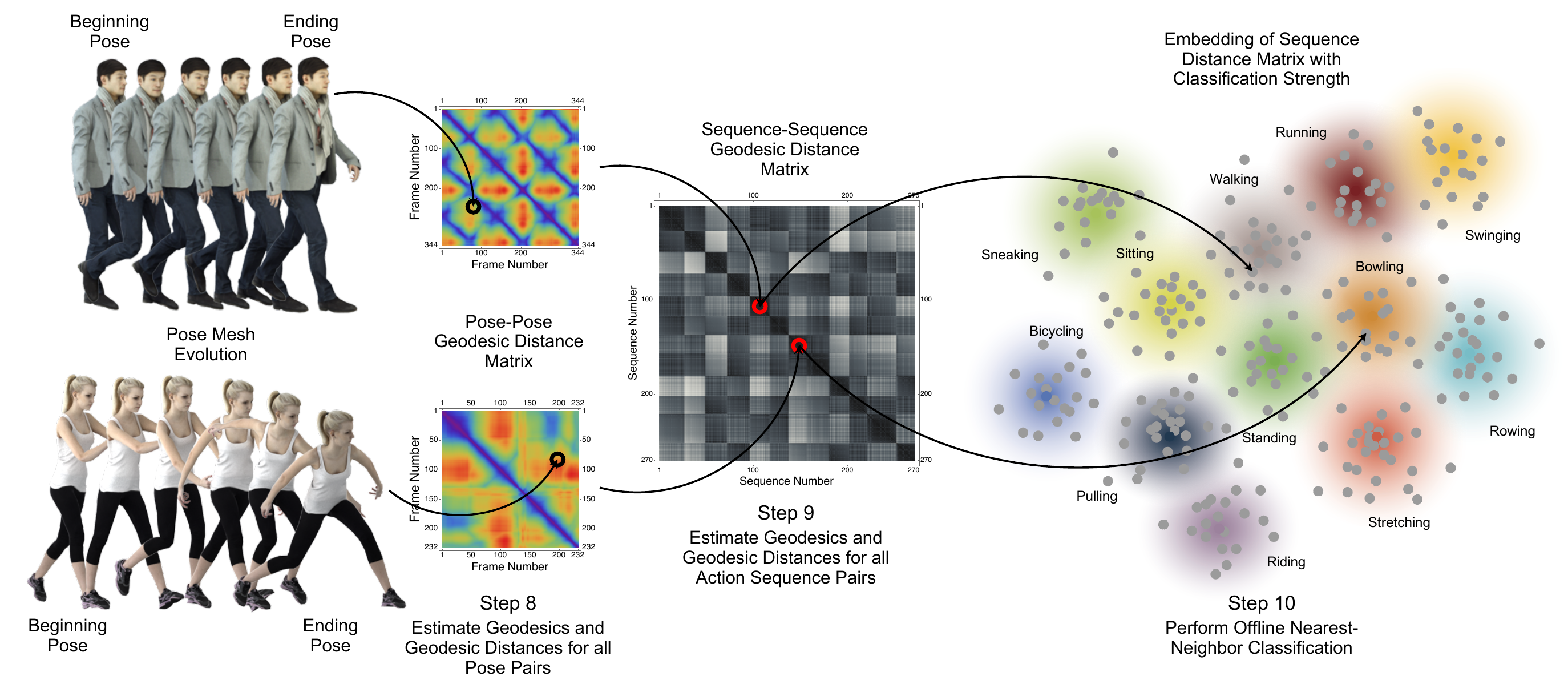}\vspace{-0.1cm}
   \caption{\fontdimen2\font=1.55pt\selectfont An overview of the motion primitive labeling process.  For each pair of poses in a given sequence, we find a geodesic between them on the orbifold of immersed surfaces.  The geodesic length between pose pairs is encoded by the pose-pose geodesic distance matrix (step 8).  This distance matrix is positive and symmetric.  The color scheme for this matrix is such that cooler colors correspond to poses with increasingly high similarity, and hence smaller distances, while warmer colors to poses with increasingly low similarity, and hence larger distances.  These matrices are useful for providing dense annotations about the motions.  Pairs of sequences can also be evaluated on a higher-dimensional orbifold of immersed space-time surfaces.  Again, geodesics are found by solving a finite-dimensional boundary-value problem using a {\sc DGM} network.  The length of the geodesic provides a pair of entries in the sequence-sequence geodesic distance matrix (step 90).   This distance matrix is positive and symmetric.  The color scheme for this matrix is such that darker colors correspond to sequence relationships with high similarity while lighter colors correspond to sequence relationships with low similarity.  This sequence-sequence distance matrix is used by a $k$-nearest-neighbor classifier to determine the appropriate action class to which the motion primitive belongs (step 10).  Here, we show where the two action sequences, from the left side of the figure, map to in the scatterplot.\vspace{-0.4cm}}
  \label{fig:fig01}
\end{figure}

Section 5 begins with a discussion of how to perform motion primitive annotation with the returned similarity matrices.  An overview is presented in \cref{fig:fig01}.  We treat the motion primitives as four-dimensional, space-time shapes.  Doing so permits compressing pairs of similarity matrices for three-dimensional kinematics sequences to a single value.  Such values describe how related two motion primitives are after sequence time warping is performed.  The latter part of the section is devoted to primitive annotation and experimental results.  We quantitatively evaluate the annotation performance when using a weighted nearest-neighbor classifier.  Our results for benchmark datasets highlight that we can achieve equal or better action annotation rates, even when using a single training example, than existing action recognition schemes and deep-neural-network approaches.  We are also able to annotate the component motions of the actions, which not all techniques can easily do.

\begin{figure}[t!]
   \hspace{0.5cm}\includegraphics[width=6.1in]{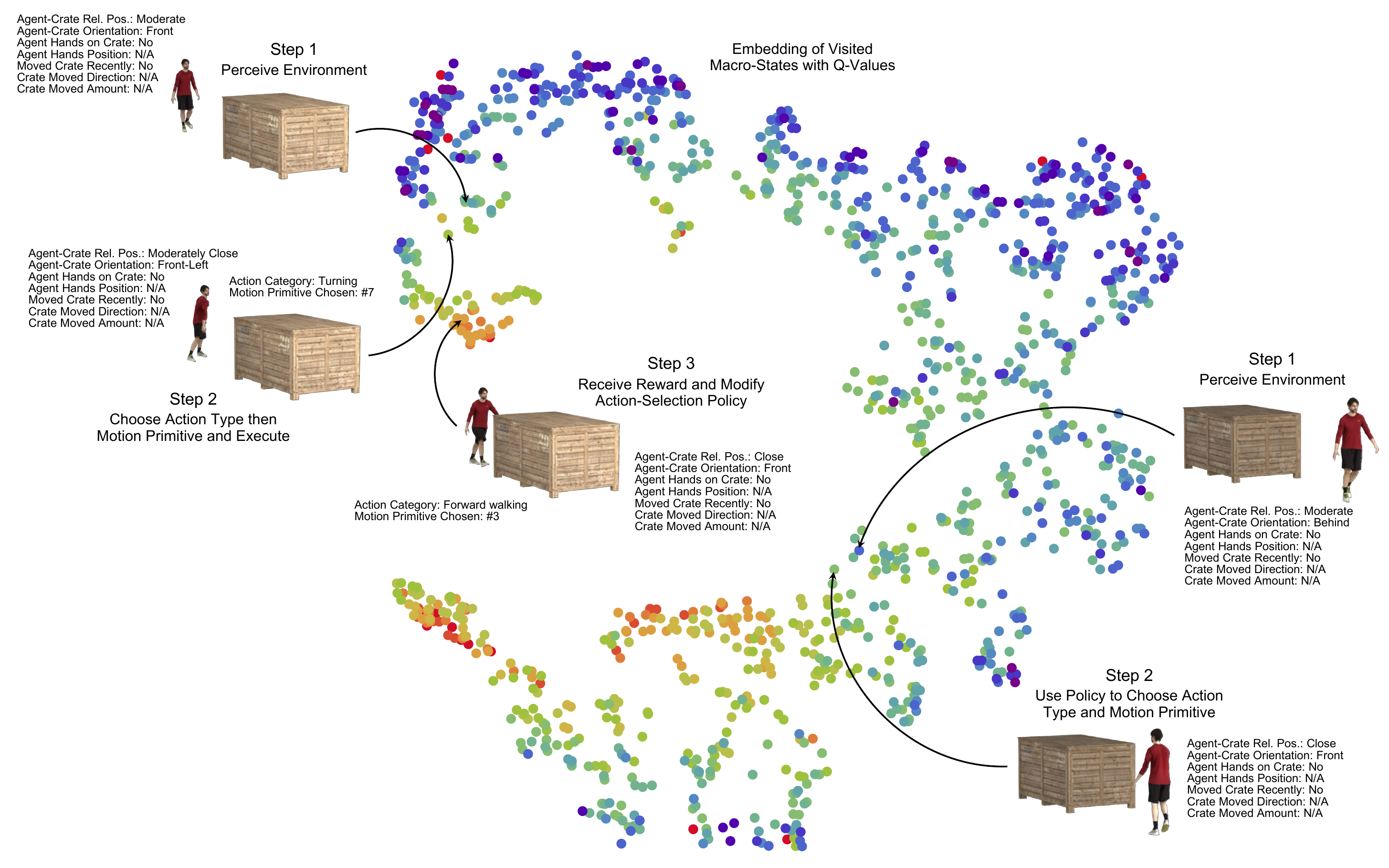}\vspace{-0.2cm}
   \caption{\fontdimen2\font=1.55pt\selectfont An overview of the reinforcement learning process with annotated primitives.  During learning, updating the policy involves three steps that are iteratively repeated until convergence (left).  First, the agent perceives the state of the environment (step 1).  Here, we have outlined the state features that correspond to the box-moving task in \cref{fig:fig6.1}.  The agent then relies on its exploration policy to determine if its current action-selection policy should be used or not.  In either case, the agent first chooses an action class and then a primitive from that class (step 2).  It then executes that primitive, experiences a sequence of state transitions, and receives a reward.  The policy is then updated based on this reward, which leads to improved agent performance over time (step 3).  For instance, instead of turning midway through a walking sequence (step 2), the agent learns to walk directly to the crate (step 3).  This leads to a state with a higher cumulative reward.  After learning is finished, the agent repeats two steps until either task completion or the environment resets (right).  It chooses an action category and corresponding motion primitive using its policy (step 1).  It then executes that primitive.\vspace{-0.4cm}}
  \label{fig:fig02}
\end{figure}

In section 6, we describe the reinforcement learning process for using annotated primitives.  We succinctly illustrate this process in \cref{fig:fig02}.  When considering manipulation tasks, annotated primitives allow for good agent behaviors to be implemented more quickly than when using non-annotated primitives.  We additionally show that the annotation rate has a profound influence on the search process.  Comparisons against classical reinforcement-learning methods highlight that our motion-trajectory-based framework can find good action-selection policies much more quickly.  This is due to an inherent search constraint imposed by using motion primitives.  That is, instead of trying to uncover the best, possibly unique, action to take for each state in a sequence of states, using motion trajectories converts the reinforcement learning problem into one of trying to find the best macro action to take over all of the states in a sequence.

\subsection*{\small{\sf{\textbf{2$\;\;\;$Literature Comparison}}}}\addtocounter{section}{1}

Much of the research on motion primitives has involved their use for path planning and control, where the aim is to construct cost-optimal paths that obey beginning and ending kinematics constraints.  In \cite{DudekD-conf2003a}, Dudek and Simhon showed how to create paths with a set of hand-picked/-sketched, non-holonomic motions.  In a related paper, {J\"{a}kel} et al.~\cite{JakelR-conf2010a} gave a probabilistic-based framework for planning with lengthy, human-derived motions.  Unlike us, \cite{DudekD-conf2003a,JakelR-conf2010a} do not employ labels for planning purposes.  We focus here on automatically labeling motion templates, that we have extracted from videos, so that primitives which are not suitable for a given task can be removed during the learning process.  As well, in contrast to works such as \cite{JakelR-conf2010a}, our motion primitives can have variable duration.

There are other works that deal with motion primitives for planning \cite{CohenBJ-conf2010a,CohenBJ-conf2011a,PowellMJ-conf2012a}.  Powell et al.~\cite{PowellMJ-conf2012a} describe so-called extended canonical human functions that characterize the locomotive modes of level-terrain walking and stair climbing.  They rely on motion capture data of human actors.  Single steps in the data are extracted and labeled through examination of heel position and acceleration data to determine instances of the feet striking and lifting from the ground.  Unlike Powell et al., we take a passive, more generally applicable approach to motion segmentation and labeling, which, for our application, involves human actions.  That is, we process single-camera video feeds and estimate pose without the need for motion capture markers.  The way in which we annotate the motions is also completely different, as we look at how the pose and position of an entity's entire body changes over time, not just a limited portion of it.  Therefore, we are able to label locomotion- and manipulation-based motions and actions.  Lastly, Cohen et al.~\cite{CohenBJ-conf2010a,CohenBJ-conf2011a} developed a means of exploiting adaptive motion primitives in search-based planning.  They manually defined the templates for a six-degree-of-freedom robotic platform.  In our work, we are interested in automatically identifying, annotating, and utilizing motion primitives with arbitrary degrees of freedom.

Some research has been performed on using motion primitives in reinforcement learning, which is a much more general and difficult problem than path planning \cite{NeumannG-conf2009a,KoberJ-conf2011a,ManschitzS-conf2014a,KroemerO-conf2015a}.  In \cite{JenkinsOC-jour2004a}, Jenkins and Mataric showed how to segment motion sequences into primitives, thereby building a motion-primitive library.  They perform segmentation using a heuristic specific to human-like kinematic motions.  As we showed in \cite{SledgeIJ-jour2019b}, this methodology segments motion sequences more poorly than our kernel-based approach.  More recently, they and other researchers used more principled methods \cite{GrollmanDH-conf2010a,NiekumS-conf2012a} to segment the data into multiple models as a way to avoid perceptual aliasing in the policy.  The authors of \cite{JenkinsOC-jour2004a,GrollmanDH-conf2010a,NiekumS-conf2012a} neither labeled the motions nor used those labels during the policy search process.  For many types of tasks, only a subset of the primitives can be considered.  Those that are ill suited for a task can be ignored, assuming that accurate action labels are available and used when forming a policy.

In \cite{KonidarisG-coll2010a}, Konidaris et al. proposed using change-point detection to segment motion trajectories into a skill chain \cite{KonidarisG-coll2009a}.  The skill chains from each trajectory are then merged to form a skill tree, which provides a path from every starting state to a goal state by executing a sequence of acquired skills.  Their segmentation approach does not, however, easily scale to the complicated three-dimensional motions that we consider here and in our previous work \cite{SledgeIJ-jour2019b}, which have nearly a hundred degrees of freedom.  Moreover, the skills that they and others learn are specified in an online fashion.  We focus here on defining skills, or macro-actions, offline.  There are advantages and disadvantages to either approach.  Learning skills in an online fashion permits solving a particular task well.  It also facilitates transfer learning to adapt the solution for similar tasks.  However, no prior knowledge is imparted by investigators about the task.  The agents may therefore waste many episodes implementing behaviors that could have been provided a priori.  In the offline case, template agent behaviors can be provided for one or many possible tasks, which can immensely speed up the trainings process.  If, however, the agent has no way of either interrupting and modifying those behaviors or defining entirely new macro-actions, then its overall task performance may suffer.  Here, we permit the agent to utilize pre-specified motion primitives and insert novel kinematics sequences between primitives.  This helps avoid performance degradation issues when the primitives cannot directly chained to solve a given task.

In \cite{SledgeIJ-jour2019b}, we proposed an approach for identifying motion primitives from long-term action sequences.  We regressed three-dimensional pose meshes from video feeds using deep networks.  We also developed a kernel-based, generalized-cross-correntropy measure to assess changes in the pose meshes over time.  We showed that this kernel measure codifies relevant relationships between natural action breakpoints and both short-term and long-term kinematics changes.  It hence facilitates motion segmentation.  Here, our aim is not motion segmentation, but rather motion annotation.  Our approach for this task is also different from that in \cite{SledgeIJ-jour2019b}.  Our proposed pose-pose similarity metric views the poses as continuous shapes.  We discretize only at the very end, when performing inference.  Our approach from \cite{SledgeIJ-jour2019b} treats the poses solely as discrete, sparse shapes, in contrast.  It is hence sensitive to the number of anchor points used for quantifying pose similarity.  Relying on only a few, well-placed anchor points may be sufficient for motion segmentation but it is often not for annotation.  Often it is necessary to consider local and global deformations of the entire body, in a continuous way, to achieve good annotation rates.  This is because different actions are sensitive to changes in position and velocity of different regions than other actions.  It is difficult to a priori specify all such regions for every action.  

\setcounter{figure}{0}
\subsection*{\small{\sf{\textbf{3$\;\;\;$Motion Primitive Similarity}}}}\addtocounter{section}{1}

Our aim is to annotate motion primitives by using details about how an entity's pose and location changes over time.  We do this using a differential-geometric approach.  It works by assuming that the poses belong to the immersion orbifold of pose kinematics.  On this orbifold, we develop a metric function that assesses the cost of deforming one pose into the other, which provides a measure of pose similarity.  When used to assess sequences of poses, this metric determines how related an unlabeled sequence is to one that is labeled, which permits transferring the annotation using a supervised classification scheme.  

\begin{figure*}[!t]
   \begin{minipage}[t]{0.65\linewidth}\vspace{0pt}%
      \includegraphics[width=4.15in]{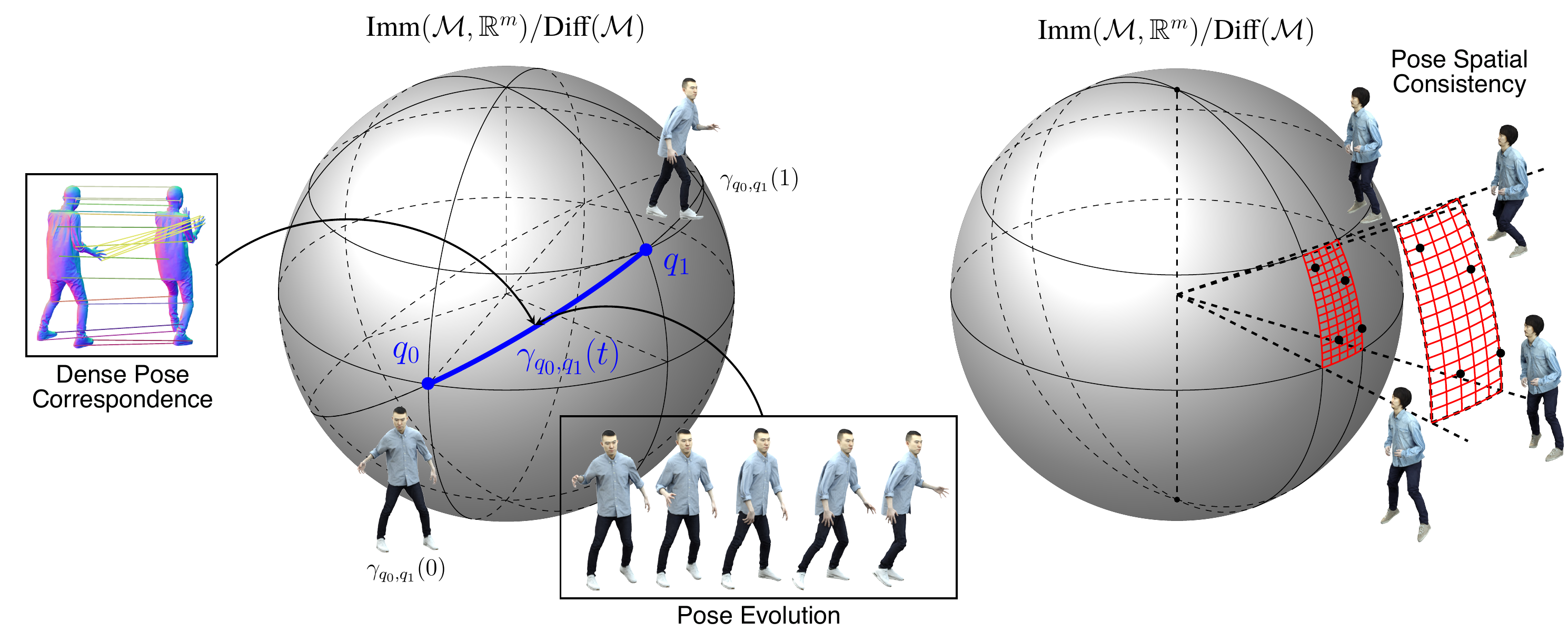}\vspace{-0.05cm}
      \captionof{figure}{\fontdimen2\font=1.55pt\selectfont An overview of our pose kinematics similarity analysis for annotation transfer.  (left) For two pose configurations, we pre-solve the pose correspondence problem.  We can then uncover geodesics, $\gamma_{q_0,q_1}(t)$, $t \!\in\! \mathbb{R}$, that connect kinematics configurations $q_0$ and $q_1$ on the orbifold $\textnormal{Imm}(\mathcal{M},\mathbb{R}^m)/\textnormal{Diff}(\mathcal{M})$.  Here, we represent the orbifold as a two-sphere for visualization purposes; the geodesic along the orbifold is denoted using the blue curve.  The length of these geodesics is taken either as a measure of pose similarity, in the case when three-dimensional pose configurations are considered, or pose-sequence similarity, in the case when a sequence of poses are considered.  From these similarity scores, the annotated sequences which differ the least from the query sequence can be found, allowing for label transfer.  (right) This approach works well because kinematics sequences that are similar can be found in local neighborhoods of the orbifold.  The amount of deformation required to go from related poses, along with related pose sequences, is minimal.}
      \label{fig:fig1}

      \includegraphics[width=4.15in]{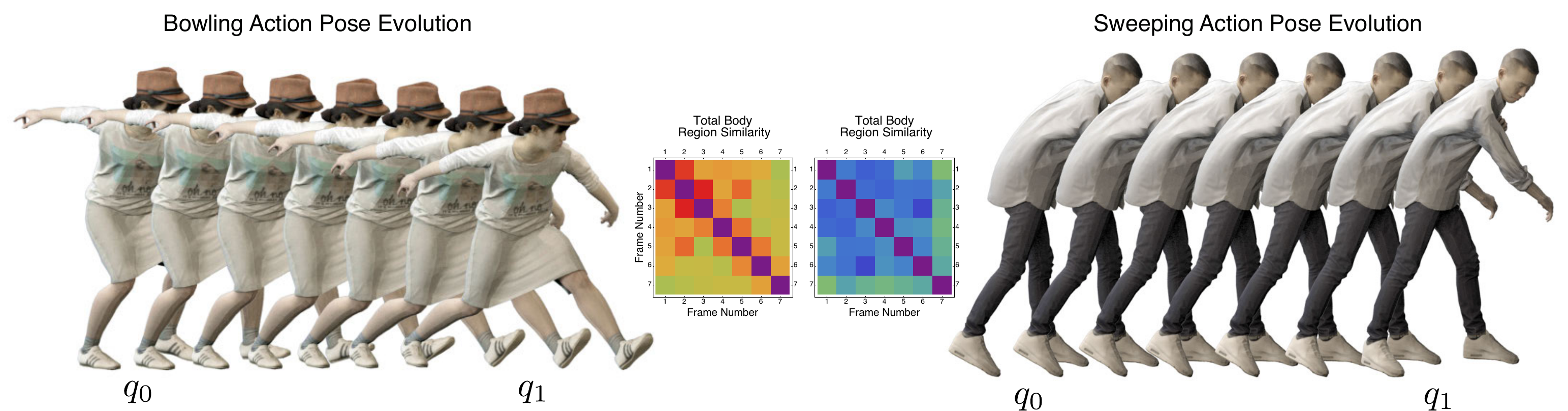}
      \captionof{figure}{\fontdimen2\font=1.55pt\selectfont Examples of pose evolutions along a geodesic for a fast-moving bowling action (left) and slow-moving sweeping action (right), along with the corresponding total-body similarity matrices for pairs of poses across different motion frames.  The beginning and ending poses are denoted, respectively, by $q_0$ and $q_1$.  Cooler colors in the matrix plots denote higher similarity, while warmer colors denote lower similarity.  The pose-similarity matrix on the left indicates large-scale changes for the quick-moving motion, as the actress' legs and arms have vastly different beginning and ending configurations; the body location also shifts.  Only small-scale changes are observed on the right, which is also captured well in the pose-similarity matrix}
      \label{fig:fig2}

   \end{minipage}\hspace{0.4cm}
   \begin{minipage}[t]{0.325\linewidth}\vspace{0pt}%
      \includegraphics[width=1.9in]{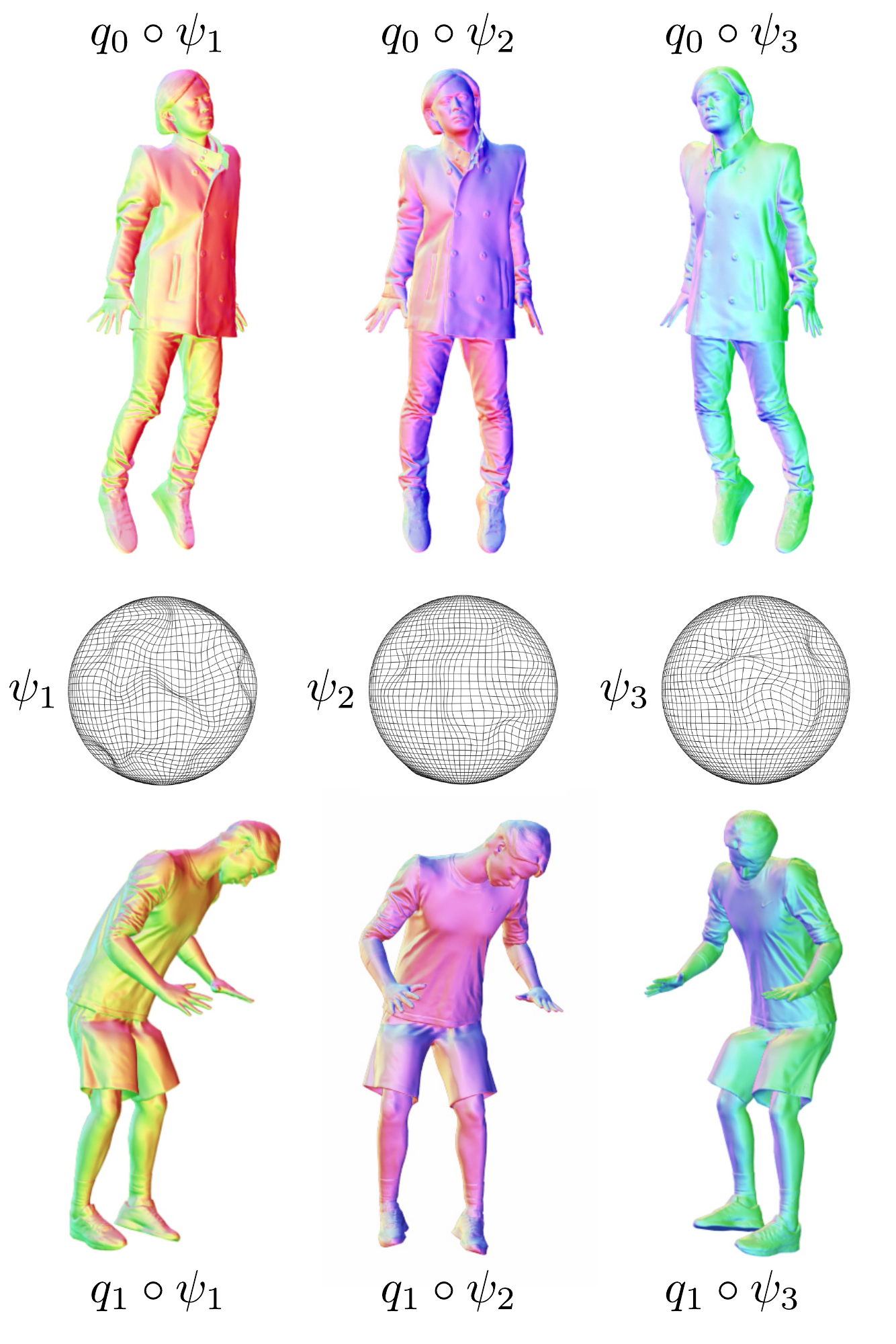}\vspace{-0.1cm}
      \captionof{figure}{\fontdimen2\font=1.55pt\selectfont Instances of two poses $q_j$ from two male actors.  These poses are presented from three viewing perspectives and with paramterizations $\psi_i \!\in\! \textnormal{Diff}(\mathcal{M})$, where $\mathcal{M} \!=\! \mathbb{S}^2$.  The parameterizations are denoted via surface coloration; here, we modified where the meshes began and ended.  The set of all diffeomorphisms is a Lie group with composition as the group operation.  The poses shown at the top and the bottom have the same shape, but unique parameterizations.  A metric that is not invariant to parameterization would result in a non-zero distance between these surfaces despite the identical shapes.  This would make comparing pose sequences difficult across multiple actors whose full-body meshes are estimated in a data-driven manner without the use of uniform templates.}
      \label{fig:fig3}
   \end{minipage}\vspace{-0.4cm}
\end{figure*}

In the rest of this section, we outline how we characterize pose similarity.  First, we formalize the space of all poses and pose sequences.  This space is a manifold of immersions \cite{KainzG-jour1984a} modulo diffeomorphisms.  Distinct sequences of variable length will be mapped to unique points on this manifold.  Next, we develop a way to gauge the distances between these points on the immersions, which will convey how related pairs of poses are to each other, in the three-dimensional case.  In the four-dimensional case, the distances will highlight the relationship between pairs of pose sequences.  In either case, the distances are insensitive to geometry-preserving transformations like translations, rotations, and scalings.  Actually computing the distance values typically involves repeatedly solving a partial differential equation of infinite dimensionality.  Here, we shortcut this process via regression implemented by a geodesic network.   

\vspace{0.15cm}{\small{\sf{\textbf{Kinematics Similarity Measure.}}}} To ensure sufficient generality, we represent motion primitives as elements of an orbifold \cite{LeeJM-book2018a}.  This orbifold is described by a connected Riemannian manifold $\mathbb{R}^m$ that is diffeomorphic to a connected and compact manifold $\mathcal{M}$.  The space of all pose sequences can be viewed as the quotient $\textnormal{Imm}(\mathcal{M},\mathbb{R}^m)/\textnormal{Diff}(\mathcal{M})$.  This orbifold is the open subset $\textnormal{Imm}(\mathcal{M},\mathbb{R}^m) \!\subset\! \mathcal{C}^\infty(\mathcal{M},\mathbb{R}^m)$ of smooth immersions of $\mathcal{M}$ in $\mathbb{R}^m$ modulo the group of smooth diffeomorphisms $\textnormal{Diff}(\mathcal{M})$ of $\mathcal{M}$.  This space is an orbifold, not a manifold, since it is locally modeled on quotients of open subsets of $\mathbb{R}^{m-1}$ by the right-action of the diffeomorphism group.

To compare either pose pairs or pairs of pose sequences, we can endow the orbifold $\textnormal{Imm}(\mathcal{M},\mathbb{R}^m)/\textnormal{Diff}(\mathcal{M})$ with a Riemannian metric.  Such a metric will characterize the similarity of three-dimensional poses and four-dimensional pose sequences to each other in terms of the length of a geodesic that connects the two kinematics configurations on the orbifold; this is illustrated in \cref{fig:fig1} with two examples shown in \cref{fig:fig2}.  This metric should possess certain properties.  For example, we would like an invariance to rotations, translations, and scalings, since the location and orientation of the camera recording the motion data should not change the action type.  As well, we would like total insensitivity to reparameterizations so that surfaces with the same shape, but different coordinate, curvature, etc., functions, have a non-zero distance, as indicated in \cref{fig:fig3}.

One possibility for a metric on $\textnormal{Imm}(\mathcal{M},\mathbb{R}^m)$ is given by,\vspace{0.1cm}
\begin{equation}
\int_{\mathcal{M}} \langle u,v \rangle_{g'} \textnormal{vol}(g)^q.
\end{equation} 
Here, $u,v \!\in\! C^\infty(\mathcal{M},\mathbb{R}^m)$ are deformation vector fields at $q$.  These tangent vectors represent the deformation vector fields of $q \!\in\! \textnormal{Imm}(\mathcal{M},\mathbb{R}^m)$.  As well, $g'$ is a fixed metric on the ambient space, and $g \!=\! q^*g'$ is the pullback metric on $\mathcal{M}$.  Part of the appeal of (3.1) is that it is the simplest metric on the immersions.  Moreover, (3.1) is not variable under reparameterizations, since the volume form $\textnormal{vol}(g)^q \!\in\! \Omega^n(\textnormal{Imm}(\mathcal{M},\mathbb{R}^m))$ reacts equivariantly to the actions of $\textnormal{Diff}(\mathcal{M})$.  Hence, it could induce a sensible distance measure on the quotient space $\textnormal{Imm}(\mathcal{M},\mathbb{R}^m)/\textnormal{Diff}(\mathcal{M})$.

\begin{figure*}[!t]
   \begin{minipage}[t]{0.245\linewidth}\vspace{0pt}%
      \hspace{0.15cm}\includegraphics[width=1.4in]{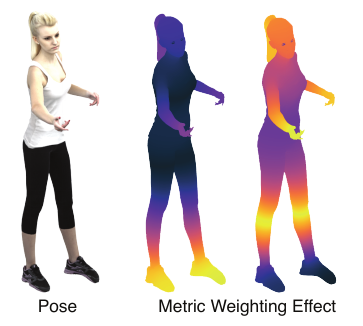}\vspace{-0.05cm}
      \captionof{figure}{\fontdimen2\font=1.55pt\selectfont Effects of two intrinsic- shape-context-based weighting terms for the reaching pose given on the left.  The left weighting term emphasizes changes in the feet and ankles more than anywhere else in the body.  The right weighting term emphasizes the knees, hands, elbow, and head.  Such terms can be incorporated into the metric to bias more for certain localized deform- ations and movements.  Doing so may aid in the recognition of actions that predominantly rely on movements of those regions.  The left-most weighting, for instance, would likely be useful for walking and jumping actions.}
      \label{fig:fig4}
   \end{minipage}\hspace{0.275cm}
   \begin{minipage}[t]{0.725\linewidth}\vspace{0pt}%
      \includegraphics[width=4.6in]{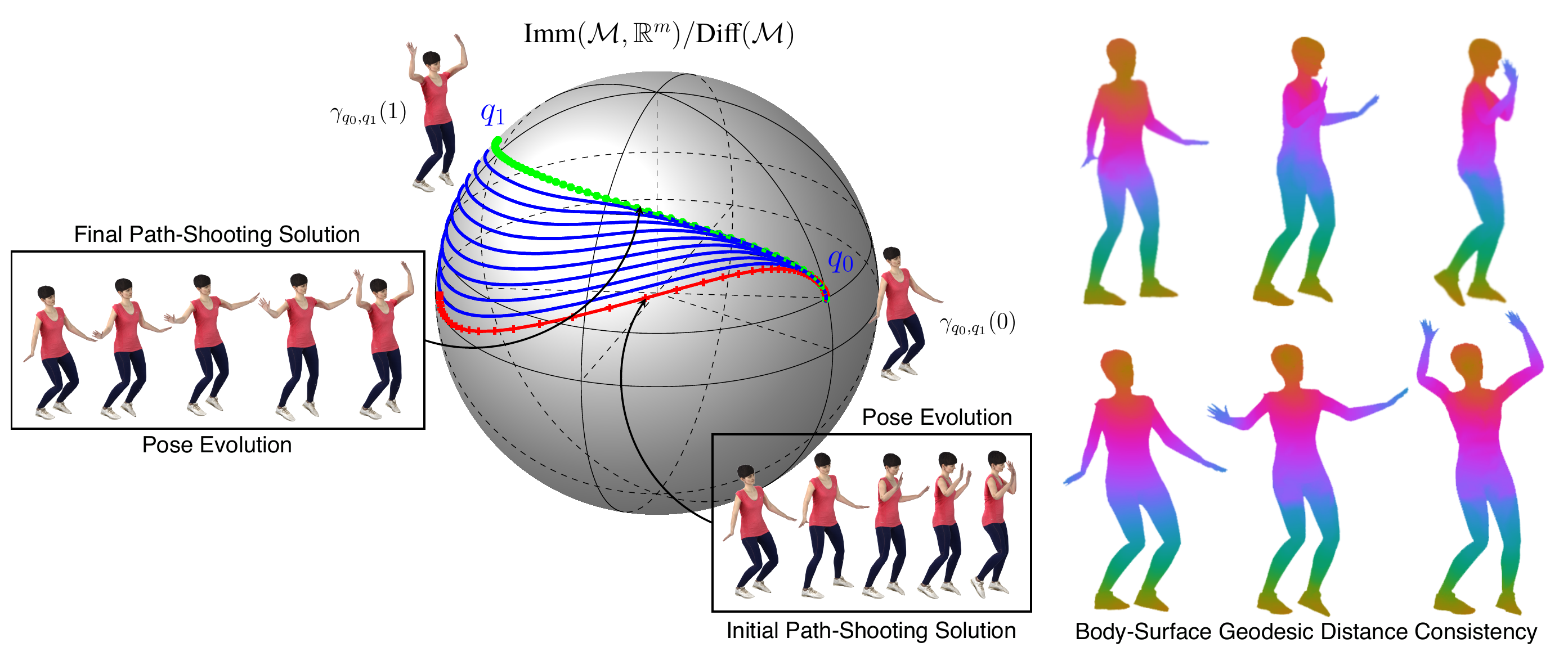}\vspace{-0.1cm} 
      \captionof{figure}{\fontdimen2\font=1.55pt\selectfont An overview of the process for finding geodesics via geodesic shooting and {\sc MDGN}s.  Given a guess for the momentum, geodesic shooting solves a boundary-value problem to specify a preliminary curve on the manifold of immersions (red line).  This curve may not terminate at the desired pose or pose sequence, though.  For example, in the case of the two target poses shown above (left), the initial solution concludes with a pose that has the legs and arms spread further apart.  We hence suitably modify the initial conditions such that the chord distance between the terminal point and the actual pose or pose sequence tends to zero.  We use a closed-form chord distance on the immersion manifold so that it is easy to compute.  This yields intermediate curves (blue lines) which more closely resemble an actual geodesic.  With enough iterations, the final pose or pose sequence should match well with the target.  Note that the intermediate poses along the shooting-based curve maintain almost the same surface area, which is illustrated by the geodesic-distance color map on the body surface (right).  If the surface area changed greatly, then we would expect the geodesic distance to be unnecessarily inflated, thereby skewing the similarities and making recognition more difficult.  Unlike geodesic shooting, {\sc MDGN}s are not iterative and can regress a geodesic (green line) that is arbitrarily close to the final solution of geodesic shooting.}
      \label{fig:fig5}
   \end{minipage}\vspace{-0.2cm}
\end{figure*}
\begin{figure*}[t!]
    \hspace{-0.4cm}\begin{tabular}{c}
       \includegraphics[height=2.1in]{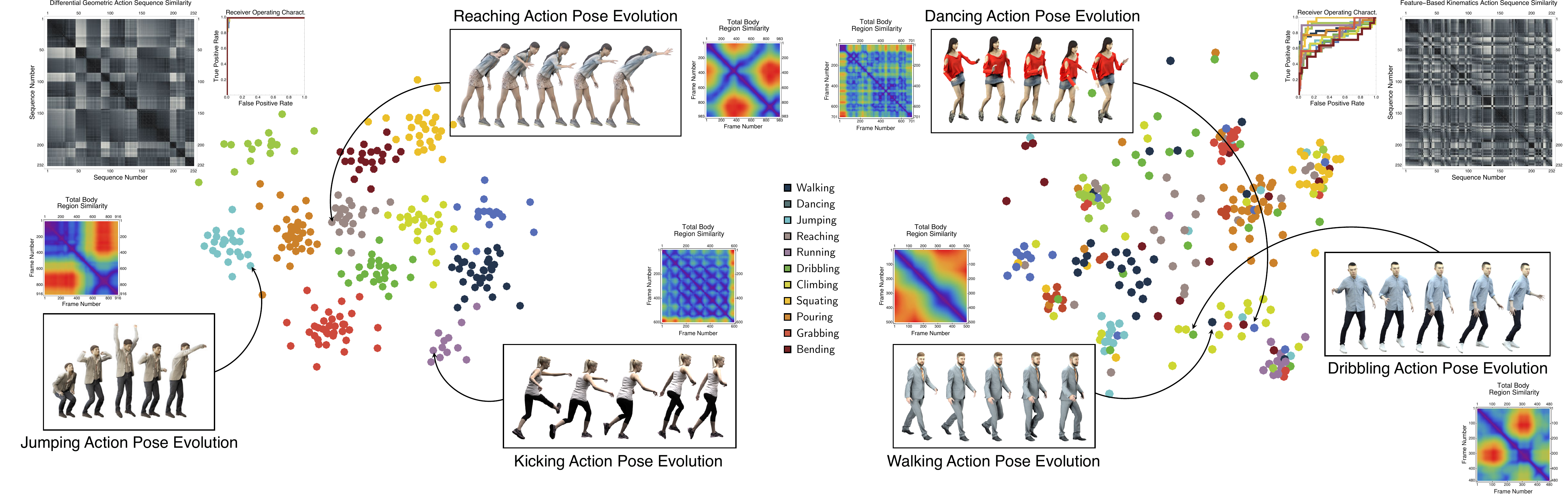}\vspace{-0.2cm}
    \end{tabular}
  \caption{\fontdimen2\font=1.55pt\selectfont A comparison of our approach with the feature-based kinematics scheme of \cite{ChenC-jour2011a}.  On left, we plot the action-sequence similarity matrix from our descriptor for 232 variable-duration action sequences that belong to one of 12 action classes.  On the right, we plot the action-sequence similarity matrix formed by using \cite{ChenC-jour2011a} for those same sequences.  For these matrices, blacker colors correspond to higher action-sequence similarity, while whiter colors correspond to lower similarity.  The similarity matrix on the left shows that 10 to 12 compact, mostly-separable clusters are naturally present, each of which correspond to a distinct action class.  High primitive annotation rates are obtained for a 1-nearest-neighbor classifier (98.1\%).  This is corroborated by the receiver operating characteristic plot.  They can be readily separated.  The similarity matrix on the right shows no dominant cluster structure.  Low primitive annotation rates are obtained for a 1-nearest-neighbor classifier (60.8\%).  This is corroborated by the receiver operating characteristic plot.  We additionally supply $t$-SNE embeddings of the pose-sequence similarities.  For some of the action sequences, we provide the corresponding pose evolution and pose-pose sequence similarity matrix.  For these matrices, high pose-to-pose similarities are depicted using increasingly cool colors, while low pose-to-pose similarities are denoted using progressively warmer colors.  Distinct patterns emerge in these pose-pose similarity plots, which indicate that our descriptor is encoding action-specific information about how the body shape changes over time.\vspace{-0.45cm}}
  \label{fig:fig6}
\end{figure*}

However, (3.1) can be shown to yield a vanishing geodesic distance: any two points in immersion manifold can hence be joined by a curve of zero length.  This metric is thus useless for our task of quantifying motion-primitive similarity, since it will return that every motion primitive is equivalent to all other motion primitives.  To ensure that the distances do not disappear, we introduce a series of $\textnormal{Diff}(\mathcal{M})$-invariant weighting functions $\varphi_i : \mathbb{R} \!\to\! \mathbb{R}_+$ that\\ \noindent depend smoothly on the immersions:
\begin{equation}
\int_{\mathcal{M}} \sum_{i=1}^p\varphi_i\!\left(\int_{\mathcal{M}}\!\textnormal{vol}(g)^q,\textnormal{tr}^{g'}\!(r),\ldots\right)\! \langle u,v \rangle_{g'} \textnormal{vol}(g)^q.
\end{equation}
Here, $r$ is a $g$-symmetric bundle mapping, with $g^{-1}r$ being the Weingarten map.  We assume that the weighting functions take as input the integral of the induced volume density, which describes the volume of the immersion; this integral is well defined due to the compactness of the manifold.  We also weight the metric by the scalar mean curvature and the position of the entity.  The use of mean curvature helps to smooth the pose evolution along the geodesic, resulting in more qualitatively reliable similarities.  Note that additional terms can be include to further bias the resulting similarities based on local and global changes, which may enhance primitive annotation performance.  Examples of local weightings are given in \cref{fig:fig4}.

For this metric to be applicable to general surfaces, the non-rigid shape correspondence problem must be solved over the diffeomorphism group \cite{KaickO-jour2011a}.  While this can be handled by either geodesic path straightening or geodesic shooting, such optimization problems are computationally prohibitive, especially for the body surfaces that we consider.  For efficiency purposes, we implement the metric using deep-Galerkin-method networks ({\sc DGMN}s) \cite{SirignanoJ-jour2018a}.  These networks approximately solve the partial differential equations that specify geodesics in the space of immersed surfaces.  We can easily use such networks to compare poses since we are only considering a single class of shapes.  If we wanted to compare arbitrary three-dimensional shapes from multiple classes, then it would be incredibly difficult to do so without copious amounts of training data.  We would be better served using either geodesic shooting or path straightening.  Those approaches generalize to arbitrary shapes readily with no training.  However, they are much slower than the DGMNs.

\begin{figure*}[!t]
   \hspace{1.45cm}\includegraphics[height=1.15in]{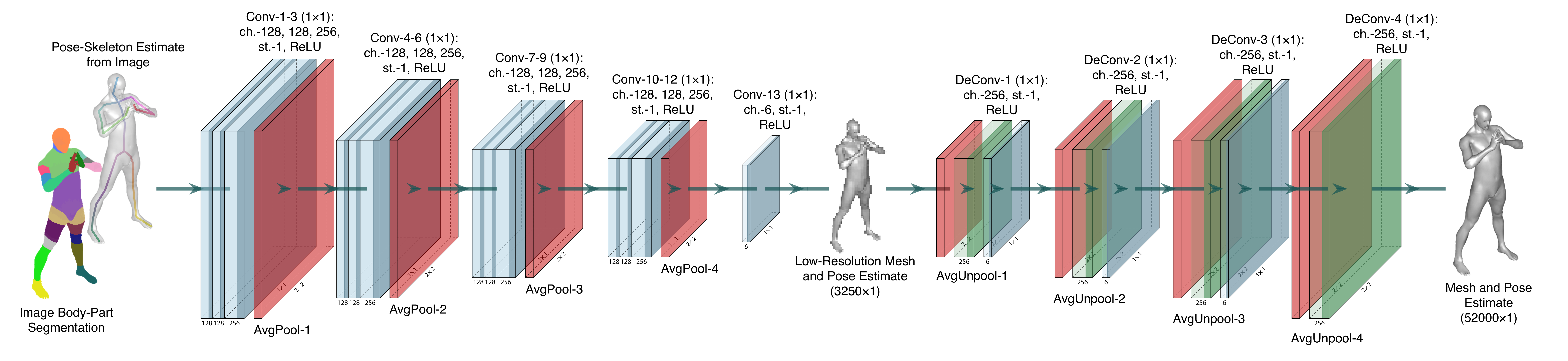}\vspace{-0.2cm}
   \caption{\fontdimen2\font=1.55pt\selectfont We rely on a small-scale {\sc DensePose} network \cite{GulerRA-conf2018a} to extract three-dimensional poses for action analysis.  Given an input image, individuals are identified and segmented.  The body parts are labeled according to an {FCN-SCP} network \cite{WangP-conf2015a}, which is not shown.  We additionally extract a two-dimensional skeleton via a {\sc UNet}, which is not shown.  Both feature maps are passed to a {\sc DensePose} network to construct a low-resolution mesh of the estimated body pose.  This mesh is progressively upsampled to create a high-resolution version.  Pairs of meshes for a time sequence are considered by a time-sensitive, graph-convolutional network to impose temporal consistency of the poses.  Here, blue-colored layers correspond to convolution, green-colored layers to deconvolution, and red-colored layers to either average pooling or average unpooling.\vspace{-0.25cm}}
  \label{fig:fig7}
\end{figure*}

\begin{figure*}[!t]
   \hspace{0.4cm}\includegraphics[width=5.9in]{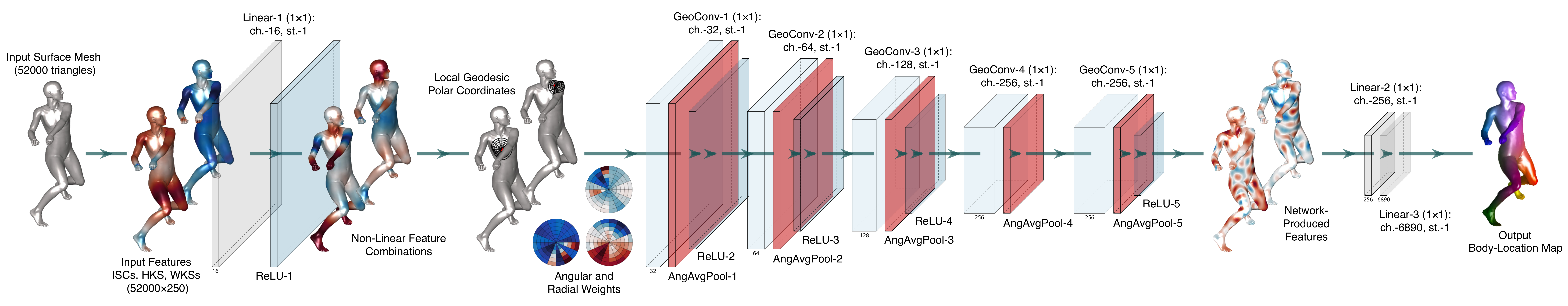}\vspace{-0.15cm}
   \caption{\fontdimen2\font=1.55pt\selectfont We rely on geodesic-based convolution networks ({\sc GCNN}s) \cite{MasciJ-conf2015a} to uncover correspondences between poses.  Given an input pose mesh, we extract a variety of dense feature descriptors, such as heat-kernel signatures (HKSs) \cite{SunJ-jour2009a,BronsteinMM-conf2010a} and wave-kernel signatures (WKSs) \cite{AubryM-conf2011a}, many of which are isometrically invariant.  Weighted, linear combinations of these features are formed, then passed through a rectified- linear activation function to form a non-linear feature response.  Local geodesic polar coordinate systems are densely formed on mesh.  The angular and radial features for each coordinate-system patch are convolved with learnable filters, average angular filtered, and then passed through a rectified-linear activation.  This process repeats multiple times to uncover progressively deep features that aid in regression of a dense body-location map for a single pose.  Pairs of poses can be related by cascading pairs of body-location maps into a small-scale convolutional graph network, which is not shown.  This resolves small-scale errors in the joint body-location map and uncover pose-pose correspondences.  Here, gray-colored layers correspond to linear operations, blue-colored layers to geodesic convolution, and red-colored layers to angular average pooling.\vspace{-0.175cm}}
  \label{fig:fig8}
\end{figure*}

\begin{figure*}[!t]
   \hspace{0.4cm}\includegraphics[width=5.95in]{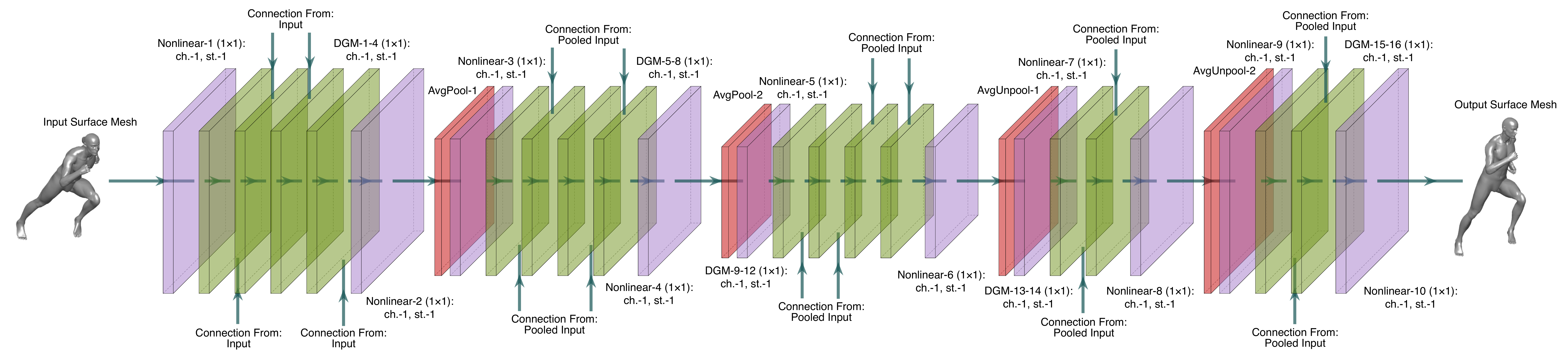}\vspace{-0.15cm}
   \caption{\fontdimen2\font=1.55pt\selectfont We use {\sc DGMNs} \cite{SirignanoJ-jour2018a} to iteratively specify geodesics.  For a given input pose shape and initial condition, this network solves a boundary-value problem to predict the corresponding pose for a unit-time geodesic on the manifold of immersed surfaces.  The initial condition is iteratively modified to ensure that the predicted pose aligns well with the target pose shape.  Solving the boundary-value problem in this way avoids the high computational overhead associated with the method of lines.  The network also addresses the problem in a meshless way, avoiding the issue of specifying local connectivity when numerically solving the partial differential equations underlying the boundary-value problem.  Here, the purple-colored layers indicate a non-linear weight and bias operation, yellow-green-colored layers correspond to a deep Galerkin model, while red-colored layers are either average pooling or average unpooling.\vspace{-0.425cm}}
  \label{fig:fig9}
\end{figure*}

The {\sc DGMN}s rely on the existence of a set of three-dimensional poses, which we extract using a modified, small-scale {\sc DensePose} network \cite{GulerRA-conf2018a}.  An overview is given in \cref{fig:fig7}.  We additionally estimate pose-pose correspondences via small-scale geodesic-convolutional neural networks ({\sc GCNNs}) \cite{MasciJ-conf2015a}, as outlined in \cref{fig:fig8}.  We post-process the registrations using manifold-based total-variation de-noising \cite{LellmannJ-conf2013a} to remove localized, small-scale discrepancies.  As well, we solve for the pose-pose rotation via Procrustes analysis and normalize the surfaces according to translation and scale.  The above metric can then be applied to the body surfaces to assess pose and pose-sequence similarity.  An illustration of this process and its comparison to geodesic shooting is provided in \cref{fig:fig5}.

This latter family of metrics has a simple interpretation.  For either a given three-dimensional pose or a four-dimensional pose sequence $q_0$, the tangent space $T_{q_0}\textnormal{Imm}(\mathcal{M},\mathbb{R}^m)$ of the immersion manifold $\textnormal{Imm}(\mathcal{M},\mathbb{R}^m)$ contains all deformations $u \!\in\! T_{q_0}\textnormal{Imm}(\mathcal{M},\mathbb{R}^m)$ of $q_0$ that are all the vector fields along $q_0$.  Given another three-\\ \noindent dimensional pose or four-dimensional pose sequence $q_1$, the tangent space $T_{q_1}\textnormal{Imm}(\mathcal{M},\mathbb{R}^m)$ of the immersions contains all deformations $v \!\in\! T_{q_1}\textnormal{Imm}(\mathcal{M},\mathbb{R}^m)$ of $q_1$ that are all the vector fields along $q_1$.  The inner product of\\ \noindent $u$ and $v$ therefore measures the amount of deformation from one pose or pose sequence to another, with respect to some continuous function $g$, as characterized by the length of the geodesic connecting the poses; see figure 3.1.  Poses and pose sequences that are similar to each other should not require much deformation.  This property helps ensure that sequence of poses with similar actions will be mostly clustered together, as depicted in \cref{fig:fig6}.  Analogous claims are not always possible for feature-based approaches, as similar poses may possess vastly different feature values.  Unrelated action sequences may thus be grouped together. 

There are multiple reasons why this measure should fare well for our primitive annotation application.  It is invariant to similitude transformations.  It can also be made invariant to affine transformations.  The recognition accuracy should thus be stable across different camera viewpoints.  An additional motivation for this measure is that it is easily extensible to the temporal case.  Instead of only relating two three-dimensional poses at a single instant, it can relate an entire sequence of poses to another sequence.  The numerical treatment of our metric has also been designed to perform a kind of dynamic time warping.  This characteristic is particularly important for label transfer, since two arbitrary motion sequences may have slight temporal differences.  By removing any short-term temporal differences, the annotation accuracy should improve.

\subsection*{\small{\sf{\textbf{4$\;\;\;$Motion Primitive Similarity Properties}}}}\addtocounter{section}{1}

Before discussing the use of our kinematics similarity measure for primitive annotation, we present some insights into the information that it provide.  We have plotted some examples in figures 4.1--4.4 for sequences containing a single, dominant action.  For these figures, we applied our metric to the three-dimensional pose sequences, which yielded a set of values that we have displayed in the total body similarity plots.  These plots characterize the similarity of each pose in an action sequence to all other poses in the same sequence.  As well, we applied our metric to local regions of the actors, such as the hands, feet, arms, and legs.  The associated plots describe how these regions move and change during the sequences.

In the remainder of this section, we qualitatively confirm that the similarities are descriptive enough for accurate action classification.  This involves demonstrating that different types of actions produce mostly unique similarity matrices.  As well, we highlight what information the similarities can impart.  We also show that the similarities are not sensitive to viewpoint changes and are not overly affected by small to moderate perturbations in the motions.

\vspace{0.15cm}{\small{\sf{\textbf{Pose Sequence Similarity Patterns.}}}} In figures 4.1--4.4, we plot the pose-pose similarities for four different actions: running while turning, ladder climbing, jumping in place, and swinging on a bar.  For each action, the total-body similarity matrices are vastly different, which suggests that our metric should be capable of distinguishing between action types.

\afterpage{\clearpage}
\begin{figure*}[p!]
\hspace{-2cm}\begin{tabular}{c}$\;$\vspace{-1.15cm}\\\includegraphics[height=10.5in]{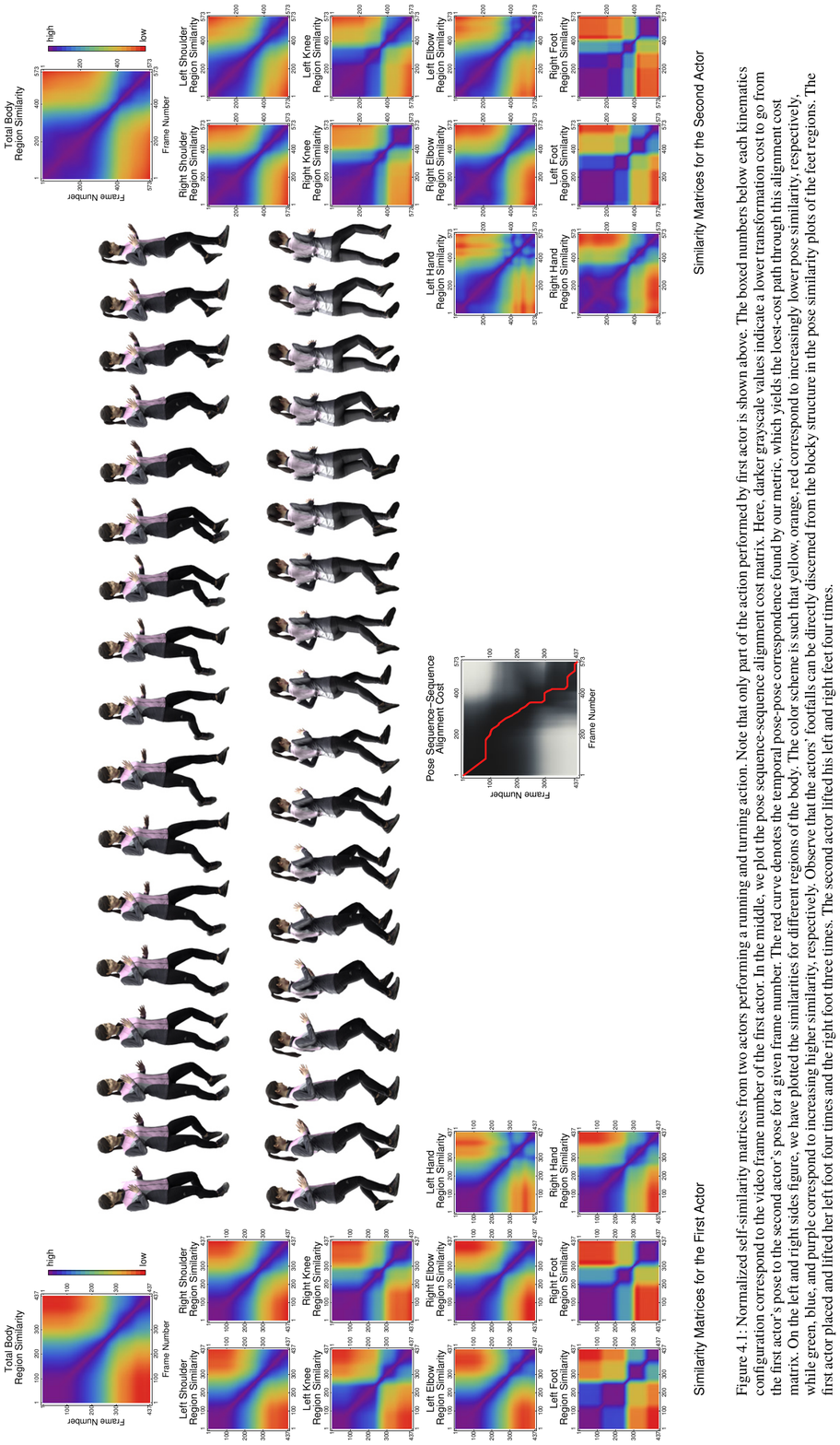}\end{tabular}
\end{figure*}

\begin{figure*}[p!]
\hspace{-2cm}\begin{tabular}{c}$\;$\vspace{-1.15cm}\\\includegraphics[height=10.5in]{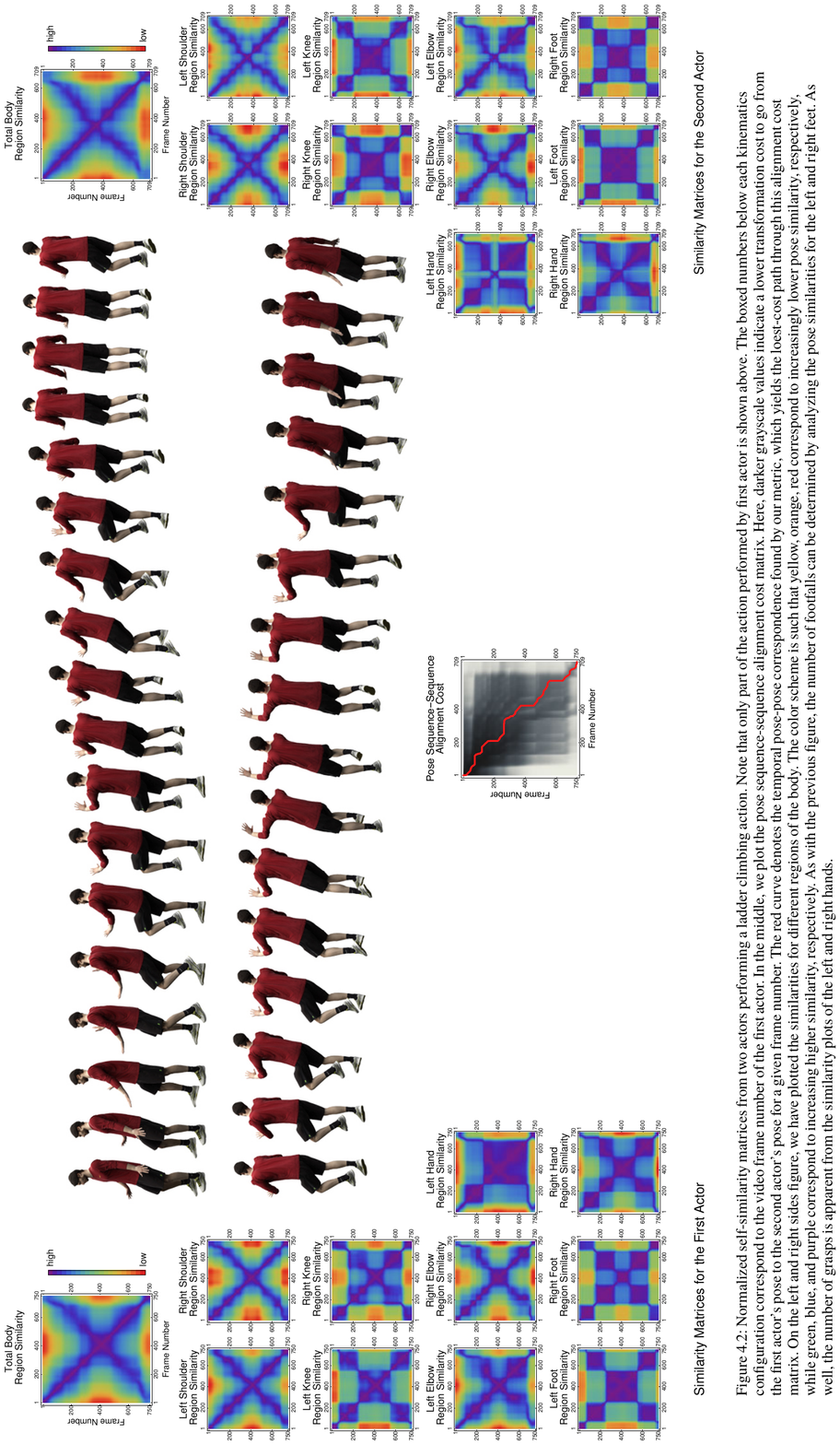}\end{tabular}
\end{figure*}

\begin{figure*}[p!]
\hspace{-2cm}\begin{tabular}{c}$\;$\vspace{-1.15cm}\\\includegraphics[height=10.5in]{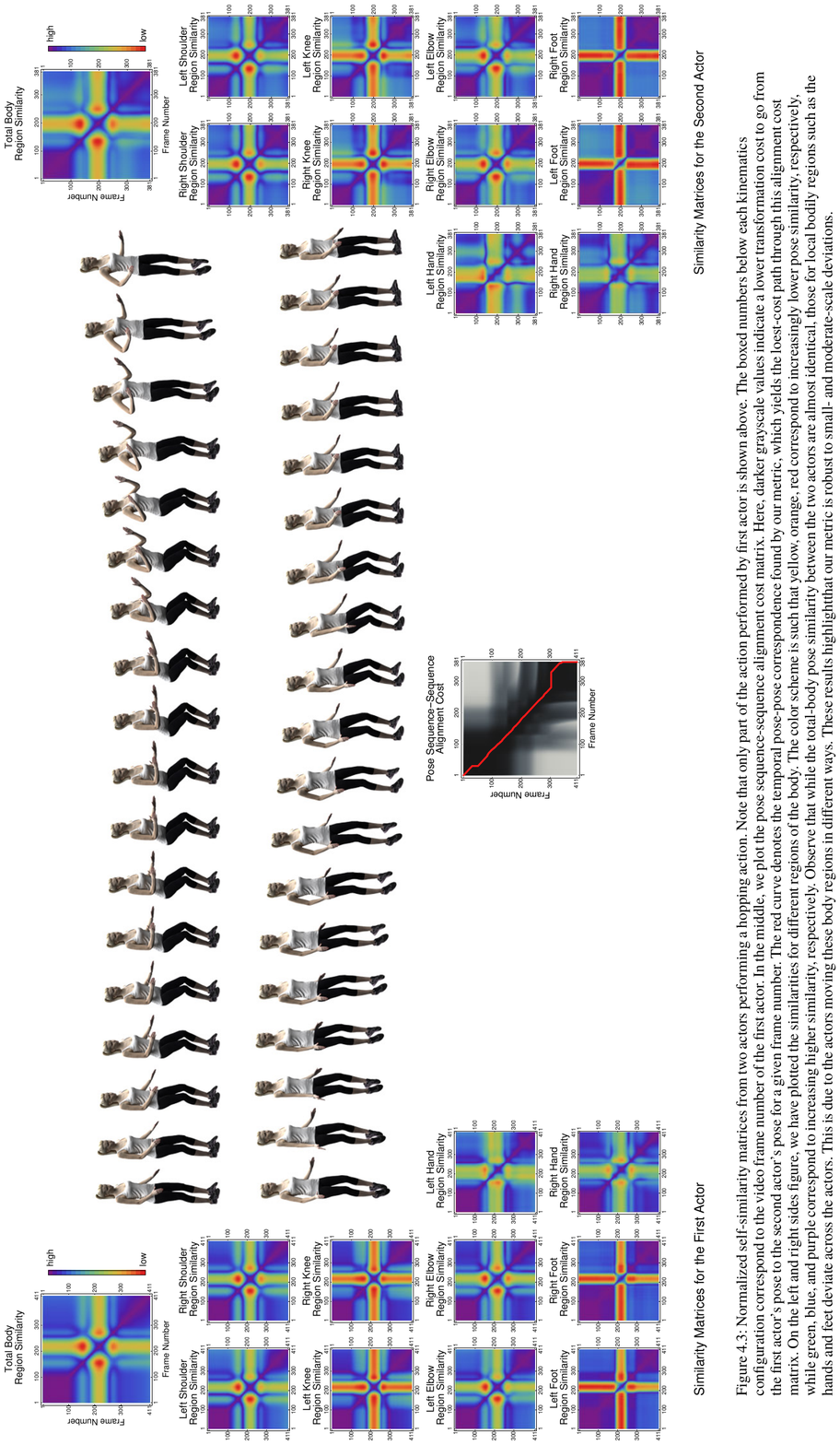}\end{tabular}
\end{figure*}

\begin{figure*}[p!]
\hspace{-2cm}\begin{tabular}{c}$\;$\vspace{-1.15cm}\\\includegraphics[height=10.5in]{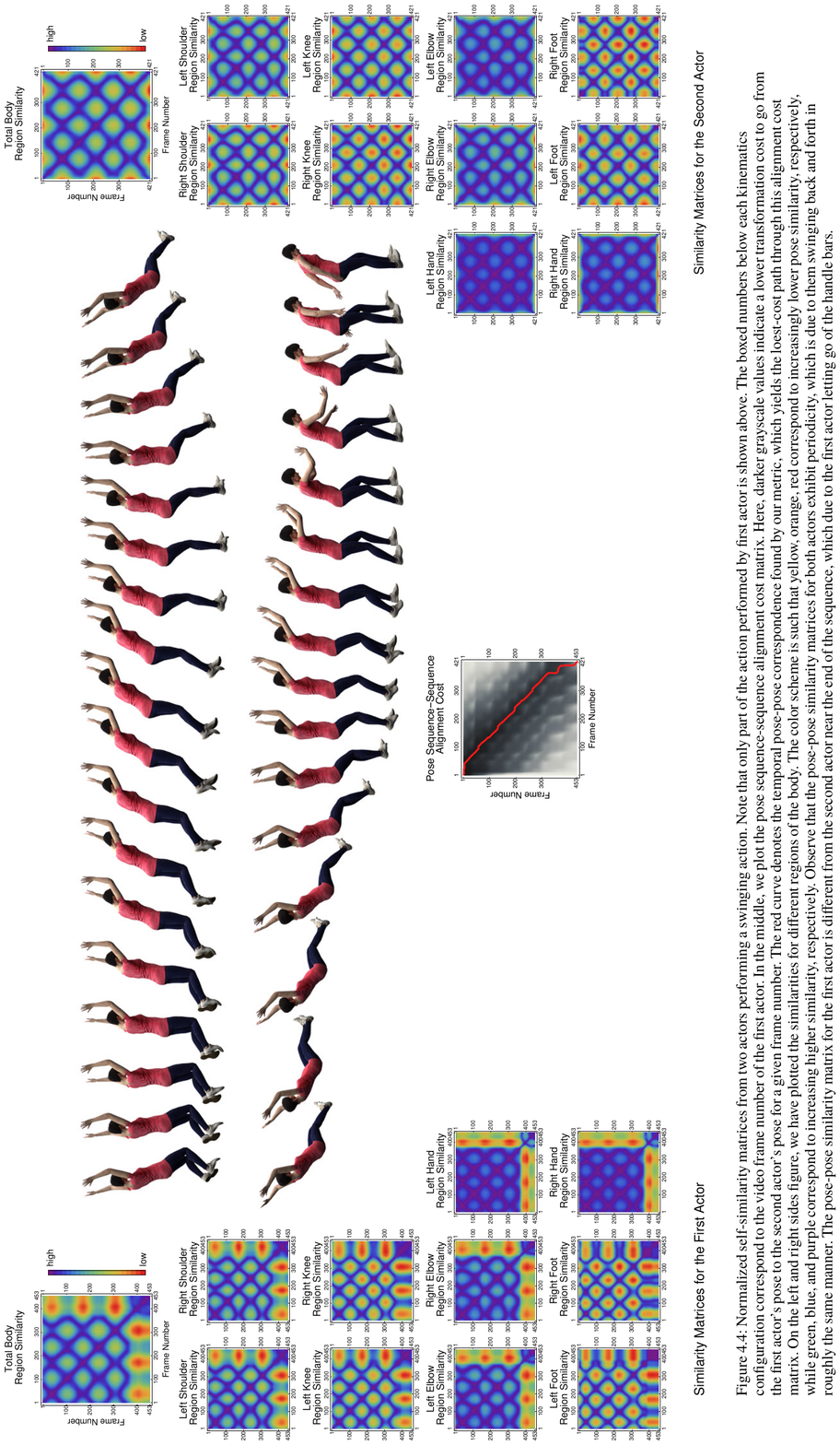}\end{tabular}
\end{figure*}

There is a wealth of action- and motion-specific patterns embedded in the pose-pose similarities.  For instance, in figures 4.1 and 4.2, the similarities highlight that the actors remained in almost constant motion for the duration of the sequence.  This is supported, for figure 4.1, by the graceful degradation in high-valued similarities (purple, blue, and green) along the main diagonal to lower-valued similarities (yellow, orange, and red) away from the diagonal in the total body plot and other body regions.  An analogous off-diagonal transition can be observed in figure 4.2.  It is, however, less gradual than in figure 4.1, since the actors briefly paused on each ladder rung.  If the actors remained stationary during most of the sequences, then we would expect to see high pose-pose similarities throughout the plots.  We would also potentially see blocks of high-similarity scores along the main diagonal of the total-body similarity matrix that were interspersed with other patterns.

Other indications that the actors for figures 4.1 and 4.2 were continuously moving are given by the blocky structure along the main diagonal for the extremity-region similarity matrices.  These high-similarity (purple and blue) blocks coincide with keeping the corresponding limb momentarily fixed.  The number of high-similarity blocks for the feet regions indicates the number of footfalls.  The red off-diagonal entries after two-thirds through the sequence indicate when the actors turned.  Note that the number of footfalls, in figure 4.1, is not immediately evident from the knee-region similarities, since the knees were bending, and hence changing position, throughout the action.  They are, however, for figure 4.2, since the actors did briefly pause.  In figure 4.2, the blocky structure in the hand and shoulder regions highlight when the actor moved up and down each ladder rung.

Thick vertical and horizontal bands of low similarities (yellow, orange and red) can symbolize that an actor was stationary for some period of time.  This occurs in figure 4.3.  For this action, the actors initially stayed relatively motionless, which is evidenced by the initial block of high similarities (blue and purple) in the upper left corner of the total body similarity plots.  The actors mainly moved only halfway through the sequence.  That is, they first flexed their knees in preparation for the jump, which can be seen by the small block of high-similarity (purple) about halfway through the sequence.  They then hopped in place, which led to the horizontal and vertical banding.  Note that there is a small period of high-similarity during the jump, which corresponds to the actors briefly hovering in place.  Afterward, the actors landed, stabilized themselves, and stood straight without moving much, which led to the other high-similarity blocks.

For certain types of actions, there are auxiliary details along the similarity matrix off-diagonals that provide insights.  Such details include the spreads and periodicities of the similarities, which allude to the abruptness of motion transitions, the cyclic nature of actions, and more.  Figure 4.4 is a good example, as there is a rhythmic pattern of diamond-shaped similarity regions (blue, green, and yellow).  The reason these structures formed is because the actors' legs swung out to the furthest point and then folded back inward multiple times.  The outward swinging of the legs produces low similarities (green and yellow), as it is both fairly distinct and short-lived.  Accompanying this diamond pattern is a repeated bisection of high similarities (purple and blue).  This bisection pattern implies that the same pose was adopted at multiple instants.  As with the footfalls, the high-similarity peaks along the main (anti-)diagonal provides clear evidence for the number of repetitions.

Some additional comments can be made about figure 4.4.  First, the hands and elbows exhibit little variation in similarity, as they remain in a relatively fixed position while gripping the bars.  What deviations do exist are due to the rotation of the palms and fingers when compensating for the swinging movement.  Secondly, near the end of the motion from the first actor, there is a period where low similarities dominate the off-diagonal.  This stems from the actor letting go from the bar and landing on the ground.

\vspace{0.15cm}{\small{\sf{\textbf{Pose Alignment and Motion Speed Insensitivity.}}}} In figures 4.1--4.4, we have plotted the temporal alignment cost between pose pairs from the first and second actors.  Darker shades in these plots indicate a lower alignment cost, while lighter shades denote the opposite.  The overlaid red curve provides the temporal correspondence between pose pairs.  If the motion sequences were of the same length and contained the same motion, then the curve would lie along the main diagonal.  Deviations from this diagonal indicate either the addition or the deletion of one or more kinematics configurations in the motion sequence due to a variety of circumstances.

For each of the actions, the temporal-correspondence curve is, more or less, along the main diagonal of the alignment-cost matrix.  This indicates that few edits are needed to align the motion sequences, despite differences in their length and the rate of speed at which the motions were performed.  As well, this alignment is possible despite that the actors had different body types, that there were local pose estimation errors, and that the motion sequences were captured from different viewpoints.  Such an outcome was expected, since the pose similarities exhibited little variation due to these factors.

\vspace{0.15cm}{\small{\sf{\textbf{Pose Similarity Discussions.}}}} Each of these pattern types indicate that our kinematics similarity metric is encoding useful details about the motions.  The total-body similarity patterns have been unique for each action considered in this section We have found this property to hold for ninety types of actions beyond those presented here.  Such observations provide qualitative assurance that our descriptor is recognizing meaningful details.  They also imply that the similarities should give rise to good classification regimes.

For figures 4.1--4.4, it can be seen that the pose-sequence similarities for each action are fairly consistent across actors.  This was despite there being relatively small differences in the motion.  One of the few exceptions is figure 4.4.  The discrepancies stem from the first actor letting go from the bar and landing on the ground; the second actor, in contrast, continued to swing.  Additionally, the physical differences in the actors were not strongly manifested in the similarities.  This implies that there should be little deviation in the labeling process for different actors and their body shapes and sizes.  As well, the camera viewpoint had little influence on the resulting similarities for these tests.  For each action in figures 4.1--4.4, the actors' poses were captured from different orientations.  If our metric was not invariant to similitudes, then we would expect to see drastic differences for each pair of sequences.

From figures 4.1--4.4, one can discern that a fair amount about the poses are redundantly recorded in multiple regions of the actor.  This initially suggests that only a few, select areas, like the hands and feet, might need to be considered by our metric when annotating the primitives.  Nevertheless, it would be incredibly shortsighted to do so, as key information that could help identify the action type may be missed.

Note that some of the information contained within the pose similarities can be derived from other sources.  The number of footfalls in a walking sequence, for example, can be obtained by processing joint angle trajectories.  Sharp transitions between one movement pattern to another can also typically be discerned from joint angle trajectories.  However, we are not calculating these similarities just to obtain such low-level details.  The pose similarities are intended to be primarily used as features for primitive annotation.  Joint angle trajectories could likewise be used for this task, but may possess too much noise to be reliable features, especially if the three-dimensional pose reconstructions are too inaccurate.  Furthermore, for a joint-angle-based annotation approach, the different joints of the body would need to be labeled when comparing sequences.  This labeling problem can be just as difficult as the problem of annotating actions and movements.  Our similarity-based approach avoids the need for such a labeling.

For each of these actions, we showed that it is possible to temporally align these sequences and perform dynamic time warping, which is important.  Such correspondences make it possible to dynamically warp one sequence to another for comparison purposes.  This property proves useful for dealing with sequences where the actors perform the same motion at different speeds.  It is also useful when a primitive contains extra motions somewhere in the sequence that another primitive does not.  A good example of this is figure 4.4.  In figure 4.4, we have the case where one actor let go of the bars while swinging and another actor who did not.  Despite this distinction, both of these sequences contain a single, predominant action: swinging.  When a measure possesses the dynamic time warping property, these types of differences can be ignored and the correct label can be transferred.

\subsection*{\small{\sf{\textbf{5$\;\;\;$Similarity-Based Primitive Annotation}}}}\addtocounter{section}{1}

In the previous section, we have qualitatively shown that the similarities from our metric have a view-stable and action-specific structure.  We have also demonstrated that the similarities are also mostly insensitive to the speed at which the motions are performed as well as the physical characteristics of the actors.

In this section, we rely on the fact that our kinematics similarity metric holds for arbitrary dimension, which allows us to construct spatio-temporal pose kinematics similarities.  The result is a single value that quantifies the similarity between two motion sequences of potentially different length.  We use these spatio-temporal similarity scores within the context of a weighted, nearest-neighbor classifier \cite{SamworthRJ-jour2012a} to perform action and motion annotation.  To help promote a high classification stability and accuracy when dealing with few training examples, we consider a bagging meta-heuristic wherever possible \cite{HallP-jour2005a}.

In the following experiments, we characterize the performance of various components of our annotation process.  First, we quantify the annotation accuracy on two publicly available datasets.  We compare our method with other action recognition procedures and consistently show higher accuracy results even when the motion speed is altered.  We demonstrate that good one-shot-learning performance can be obtained with our approach.  Next, we show that we are able to find temporal correspondences between pairs of action sequences.  Lastly, we analyze the viewpoint and pose-noise insensitivity of our metric on a third publicly available dataset.

\subsection*{\small{\sf{\textbf{5.1$\;\;\;$Primitive Annotation Preliminaries}}}}

We consider three benchmark datasets for our experiments.  For each, we automatically extracted the three-dimensional pose from the action-sequence videos using the deep-neural-network approach from \cite{GulerRA-conf2018a}.  We then applied our motion primitive identification algorithm from \cite{SledgeIJ-jour2019b}, which commonly returned only a single primitive for each video.  Multiple primitives were returned only when the video sequences contained a repeated action.  In these cases, the primitives were virtually identical and the same label was assigned to each of them.

For most of our experiments, we randomly split the motion primitives from each dataset into training and test sets.  The results that we report for our approach are the average test-set performance across 100 Monte Carlo trials.  Five nearest neighbors were used to transfer the action and motion labels, as we found this amount to work well across all of our experiments.  When performing comparisons, we list the recognition rates provided by other researchers, who often used substantially more training examples.  For the one-shot learning experiments, we randomly considered a single training sample for each class and report the average performance across twenty different runs.

\subsection*{\small{\sf{\textbf{5.2$\;\;\;$Primitive Annotation Results and Discussions}}}}

{\small{\sf{\textbf{Baseline Performance and Motion Speed Insensitivity Results.}}}} We first quantify the baseline performance of approach, which we do using two publicly-available, benchmark video datasets.  The first is the Weizmann Institute dataset, which contains low-resolution, single-view feeds of actors performing seven types of actions: bending, jumping, jumping in place, running, skipping, walking, and waving.

Although the Weizmann videos are of low resolution, we were able to attain a 100\% correct action and motion classification rating for all categories using multi-fold cross-validation.  When each class contained only a single training example, we obtained the same classification rate.  These results compare favorably with the feature-based action recognition findings of Gorelick et al.~\cite{BlankM-conf2005a} (97.5\%), Ali et al.~\cite{AliS-conf2007a} (92.6\%), Jhuang et al.~\cite{JhuangH-conf2007a} (98.8\%), Liu et al.~\cite{LiuJ-conf2008a} (90\%), Schindler and van Gool~\cite{SchindlerK-conf2008a} (100\%), Fathi and Mori~\cite{FathiA-conf2008a} (100\%), Zhang et al.~\cite{ZhangZ-conf2008a} (92.9\%), and Bregonzio, et al.~\cite{BregonzioM-conf2009a} (96.6\%).  The deep neural network approaches of Ji et al. \cite{JiS-jour2013a} did poorly (90.2\%) due to the lack of sufficient quantities of training data.  Note that these referenced techniques are not primitive annotation methods; they solely recognize actions.  Primitive annotation requires not only discerning the type of action being performed, e.g., `bending', but also tagging the body movements, e.g., `left and right knees flexing' and `torso rotating forward'.

To further quantify the performance, we conducted trials on the KTH dataset.  The KTH dataset contains six types of actions: boxing, waving, clapping, walking, jogging, and running.  These actions were performed repeatedly by 25 actors in indoor and outdoor settings.  Overall, the KTH data is more challenging than the Weizmann data, due to the large variations in the body shape, view angles, scales, and appearance.  Despite these challenges, the action (99.4\%) and motion (99.7\%) classification accuracy dropped by only a small amount.  On the other hand, the feature-based action recognition techniques of Schindler and van Gool \cite{SchindlerK-conf2008a} (92.7\%), Laptev et al.~\cite{LaptevI-conf2008a} (91.8\%), Kim and Cipolla \cite{KimT-jour2009a} (95.3\%), Rapantzikos et al.~\cite{RapantzikosK-conf2009a} (88.3\%), Ali and Shah \cite{AliS-jour2010a} (87.7\%), and Seo and Milanfar \cite{SeoHJ-jour2011a} (95.1\%) had more difficulties.  Deep neural network schemes \cite{JiS-jour2013a} again did poorly (89.7\%) except in the case where they were pre-trained on data from other sources (94.1\%).  When relying on only a single training example for each class, our action recognition rate is superior to those listed (98.6\%).  The motion classification rate is also still quite high (99.1\%).

We additionally used the Weizmann and KTH datasets to illustrate that our metric is insensitive to the speed at which an action is performed.  Neither speeding up nor slowing down the training-set motion sequences by one-and-a-half times impacted the annotation performance.  Likewise, the motion classification accuracy did not change.  Other techniques, such as those based on deep neural networks \cite{JiS-jour2013a} (80.5\%) and recurrent networks \cite{DuY-conf2015a} (87.4\%), had substantial difficulties recognizing the actions in these circumstances, due to the lack of relevant training data.  The performance from the feature-based approaches by Schindler and van Gool (81.7\%), Ali and Shah (77.2\%), and Seo and Milanfar (83.5\%) also dropped when the motion sequences were either sped up or slowed down by one-and-a-half times.  Altering the motion rate-of-speed beyond one-and-a-half times led to increasingly worse performance for these other approaches.

\vspace{0.15cm}{\small{\sf{\textbf{View Insensitivity and Noise Robustness Results.}}}}  Next, we test impartiality of our descriptor to multiple camera views along with its ability to overcome noise.  To do so, we used the EPFL IXMAS benchmark dataset, which contains thirteen types of actions carried out by ten actors from five vantage points.  For our testing protocol, we considered three setups: cross camera, same camera, and any-to-any.  By cross camera, we mean that our framework was trained on action sequences from one view and tested on those from a camera that was oriented opposite to the first.  By any-to-any, we mean that our framework was trained on sequences from one camera and tested on those from one of the remaining four cameras.

Excluding the IXMAS top-down camera footage, the action recognition rates for cross camera, same camera, and any-to-any setups were 97.8\%, 98.9\%, and 97.6\%, respectively, when training on three actors and testing on the remainder.  The motion annotation rates were 98.6\%, 99.5\%, and 98.9\%, respectively.  As with the above experiments, one-shot classification did not degrade the annotation quality by much (96.2\%, 97.9\%, 97.1\%).  Inserting small amounts of random kinematics noise into the sequences also did little to disrupt either the action (97.1\%, 98.4\%, 97.3\%) or the motion (98.2\%, 98.9\%, 98.3\%) recognition accuracy; only larger amounts had any meaningful impact.  The base recognition rates of our approach were nearly one and a half times the action recognition rates of Laptev et al.~\cite{LaptevI-conf2008a} (61.8\%, 80.6\%, 64.3\%), Farhadi and Tabrizi \cite{FarhadiA-conf2008a} (58.1\%, 68.8\%, 60.3\%), and Junejo et al.~\cite{JunejoIM-jour2011a} (61.8\%, 74\%, 64.3\%) despite their use of significantly more training samples for each action.  Perturbing the pose kinematics with noise further degraded the performance.  As well, we found that deep neural networks underperformed \cite{JiS-jour2013a} (82.3\%, 89.0\%, 77.5\%) compared to our approach.

One of the advantages of our differential-geometric measure is it gauges the movement changes of the entire entity when assessing pose and pose-sequence similarity.  It can also be used to analyze specific body regions.  While targeting a small number of regions can significantly reduce the similarity-calculation time, it often adversely impacts performance, depending on the action being considered.  For instance, when just considering the hands and feet, the performance for certain actions, like throwing (99.5\%, 99.7\%, 99.3\%) did not decrease much (98.2\%, 99.1\%, 98.5\%), since much of the informative motion content was centralized in these regions; including more regions lead to a negligible improvement depending on the camera orientation.  For actions like sitting down, the average recognition rate degraded markedly when only considering these two regions (72.3\%, 77.5\%, 70.2\%).  In these cases, it was necessary to include the motion changes in the knees and elbows to accurately transfer labels (96.4\%, 97.7\%, 95.0\%).  For other actions, considering the hips, neck, and head can prove crucial.

Each of the action sequences in the IXMAS dataset was not temporally synchronized, as the actions took a variable amount of time to complete.  Some degree of dynamic time warping was therefore necessary to handle the movement discrepancies and ensure accurate annotation.  As we noted above, our similarity measure is able to automatically discern a suitable alignment and corresponding non-linear sequence warping.  We found that the warping favored by our kinematics similarity metric improved upon the solutions of alignment techniques by Rao et al. \cite{RaoC-conf2003a} (25.3\%), Ukrainitz and Irani \cite{UkrainitzY-conf2006a} (37.5\%), and Singh et al. \cite{SinghM-conf2008a} (27.1\%).  It also performed well when spurious intermediate poses were introduced into the sequence.

\vspace{0.15cm}{\small{\sf{\textbf{Primitive Annotation Discussions.}}}} With these experiments, we have shown that our kinematics similarity measure is robust under many conditions and hence achieves a high action and motion annotation rate that should prove useful for reinforcement-learning purposes.  This high accuracy is due to a variety of reasons.  Foremost, as predicted by theory, the metric is invariant to similitudes and can often correctly label actions and motions from different camera viewpoints.  It can also uncover near-optimal spatio-temporal correspondences, which allows for reliably label transfer between either slightly different pose sequences or those with actions that take are completed at a different rate of speed.  Moreover, the similarities are mostly insensitive to the actors' physical characteristics.

Relying on a global, three-dimensional pose reconstruction process versus a local, feature-based representation of the motions derived from histograms of oriented gradients \cite{JunejoIM-jour2011a,LaptevI-conf2008a} or optical flow histograms \cite{LaptevI-conf2008a,FarhadiA-conf2008a} also aided in performance.  Although it can be challenging to arrive at a good three-dimensional reconstruction, its utility appears to be greater than independently tracking body markers over time.  For example, marker points can sometimes be lost, due to body part occlusions, or improperly tracked, due to changes in lighting, shadows, or quick motions.  All of these effects contribute to the uncertainty of tagging an action.  In contrast, the manner in which we reconstruct the poses allows us to address these problems and others like (self-)occlusion \cite{UrtasunR-conf2006a} and limb following \cite{YangY-conf2011a,TaylorJ-conf2012a} well.

Although we proved that our metric is invariant to certain types of camera transforms, we did not achieve perfect classification accuracy on this data.  There are a few reasons for this.  Foremost, for some of the camera viewpoints, there can be major pose reconstruction errors.  For example, in using a top-down camera view, many of the limbs may be obscured, which makes it difficult to estimate their locations.  The similarities are also not implicitly invariant to perspective transformations.  As well, there can be different types of actions that strongly resemble each other.  Jogging and running can appear highly similar in certain situations, for example.  If enough training examples are not provided, then the classifier may not accurately distinguish between the actions.

We have additionally demonstrated that focusing on a few limbs is no substitute for processing the entire body when calculating curvature-weighted, space-time kinematics similarities.  Toward this end, we examined three cases: inspecting only the hands and feet, additionally including the elbows and knees, and including the hips and head.  We discovered that the action classification accuracy rose for the benchmark datasets as more areas were inspected.  This outcome is not surprising.  Many actions rely on cues from multiple parts of the body and are not localized to just the hands, feet, and surrounding regions.  This only occurred, though, when utilizing local shape curvature.  Without it, we witnessed a severe drop in performance for all three datasets, which indicates that it is essential for characterizing how the body is deforming at each joint and hence for detecting different actions.

Our pose similarity measure naturally uncovers temporal correspondences.  This property facilitates accurately recognizing actions and motions, almost independent of their rate of speed, provided that the kinematics can be sufficiently resolved.  As well, it can tolerate spurious kinematics configurations in a sequence well.  Certain types of feature-based approaches have difficulties realizing this behavior, since the chosen features can be highly sensitive to temporal shifts and discontinuities.

In our experiments, we highlighted that our kinematics similarities allow for accurate annotations with either many or just one training example.  Using a sole training example has advantages.  One advantage is that it reduces the amount of labeled training data that needs to be supplied, which permits constructing large-scale motion-primitive libraries prior to reinforcement learning.  Including additional observations can, however, provide an increasingly complete view of the actions' statistics and thus yield a higher overall performance, as we noted above.  For the considered benchmark datasets, though, the difference in classification rates in the one- and multi-shot learning scenario was fairly low.  We attribute this outcome to the expressiveness of our descriptor and its ability to tolerate pose-sequence discrepancies.  Other action recognition schemes, especially those that are deep-learning-based, often do not perform well with just one or even a few labeled examples.  They hence would be a poor fit for our reinforcement learning application because investigators might need to hand-label several hundred or thousand motion sequences to achieve a reasonable level of performance.

\subsection*{\small{\sf{\textbf{6$\;\;\;$Reinforcement Learning with Annotated Primitives}}}}\addtocounter{section}{1}

In the previous section, we highlighted that the motion-primitive similarities achieve high annotation performance on several benchmark datasets.  Quantitatively, we have shown that it is mostly speed and view-point invariant.  It is also robust to temporal disruptions in the motion sequences and rather insensitive to pose-estimation noise, regardless of the type of action being performed.  It hence should prove highly effective for labeling a wide variety of primitives prior to their use for reinforcement learning.

In this section, we showcase the effectiveness of annotated motion primitives for facilitating manipulation tasks within a semi-Markov-decision-process abstraction.  We demonstrate that semi-Markov $Q$-learning with annotated primitives leads to better policies more quickly than skill chaining and standard semi-Markov $Q$-learning either with or without non-annotated primitives.  We additionally quantify how the primitive annotation accuracy influences the policy learning rate and solution quality.

\subsection*{\small{\sf{\textbf{6.1$\;\;\;$Reinforcement Learning Preliminaries}}}}

For our reinforcement-learning experiments, we consider two object-manipulation tasks that take place in the Unreal Engine 4 environment.  For the first task, a humanoid agent must discern how to efficiently move a heavy crate from a random starting location in an obstacle-free domain to a pre-specified goal location.  In the second task, the agent should determine how to wipe sections of a flat surface with a sponge.  The size and position of the sections to be wiped randomly evolve as the agent completes more of the task, as does the difficulty in clearing a section.

\vspace{0.15cm}{\small{\sf{\textbf{Task Reward Structure.}}}} We utilize the following reward structure for these two problems.  For the first, the agent receives a small reward ($+$5) for maneuvering the crate a set distance closer to the goal location and a small penalty ($-$20) either for moving it a set distance away from the goal or for the agent moving away from the crate.  The agent is also penalized ($-$1) for every action that it takes, which discourages forming policies with unnecessary behaviors.  Moving the crate to the goal location results in a high reward ($+$1000).  Doing so within a certain number of action choices leads to an additional reward ($+$1500), which helps promote efficient strategies.  For the second task, the agent receives a small reward ($+$100) for finding the sponge and for wiping a set area of the surface ($+$10).  Harder-to-wipe sections, which require two or three passes to clear, lead to larger rewards ($+$150).  Taking any action is again penalized ($-$1); moving away from the surface ($-$50) or dropping the sponge ($-$250) are too.  Clearing the entire surface yields a high score ($+$1000), as does finishing the task within a set time ($+$1500).

The agent has a pre-specified number of actions that it can take before the environment resets.  When not using motion primitives, it is twenty-thousand arbitrary kinematics choices.  When using motion primitives it is a hundred macro-actions and up to five hundred arbitrary kinematics configurations.  The agent is given three tries per episode to complete the task.

\vspace{0.15cm}{\small{\sf{\textbf{Task State-Action Space.}}}} We consider different action spaces for both tasks.  The first allows the agent to specify one of approximately ninety-thousand unique bodily kinematics configurations.  This space corresponds to the case of conventional reinforcement learning where no prior knowledge is available about the tasks to constrain the action-selection process.  The second space relies on the first and also uses variable-duration motion primitives as the action choices.  When dealing with labeled motion primitives, the agent either specifies a kinematics configuration or first selects the action class and then the primitive to be executed.  The agent has access to approximately ten-thousand motion primitives across over a hundred action classes.  When labels are not available, the agent simply specifies either which primitive is to be performed or which kinematics configuration should be targeted.  In each case, the ending and beginning kinematics are interpolated across a state series to yield smooth movement trajectories.

We extracted the motion primitives by processing high-resolution, curated video feeds of one or more actors carrying out various tasks that involved combinations of manipulation and locomotion.  Our temporal segmentation approach from \cite{SledgeIJ-jour2019b} was employed to identify primitives, and our labeling scheme that we developed here was used to annotate them.  We achieved high action (98.7\%) and motion (98.3\%) recognition rates when relying on five labeled samples per action class.

A variety of state features are used to describe the domains.  We consider nine manually-defined features for the crate-moving task.  These include the relative position and orientation of the agent with respect to both the crate and the goal.  They also include if the agent's hands are on the crate and where they are positioned, if it has moved the crate recently and, if so, in what direction and by how much during that time.  For the wall-wiping scenario, fifteen manually-defined features are extracted.  We assess the relative position and orientation of the agent with respect to the surface and the nearest region that needs to be cleaned.  The agent is also made aware of the dimensions of the regions to be cleaned, if they require multiple passes, if and in what hand it is holding a sponge and, if not, where that sponge can be found in the domain.  As well, we determine where the agent's hands are relative to the nearest section to be cleaned and if it has recently wiped near that location.  When motion primitives are employed, we augment the feature sets for both tasks with the action and motion annotations for the last-used primitive.  As well, for both tasks, we include an additional ten attributes obtained from deep convolutional autoencoder networks that operate on the current image frame and on the past five frames, all of which are from the agent's perspective.

\vspace{0.15cm}{\small{\sf{\textbf{Reinforcement Learning Protocols.}}}} We used semi-Markov-decision-process-based $Q$-learning \cite{BradtkeSJ-coll1994a} with value-of-information-based exploration \cite{SledgeIJ-jour2017b,SledgeIJ-jour2017c} and experience replay \cite{LinLJ-jour1992a} to formulate action-selection policies.  Locally-linear experience generalization was employed to interpolate value-function magnitudes to state-action pairs.  Multiple parameters must be set for these approaches.  For semi-Markov $Q$-learning, we relied on an inverse polynomial decay schedule for the learning rate, from a value of 1.0 to 0.0001, to facilitate polynomial-rate policy convergence \cite{EvenDarE-jour2003a}.  A discount factor of 0.8 was used to preempt slow convergence \cite{SzepesvariC-coll1997a}.  We applied our path-following approach \cite{SledgeIJ-jour2019a} to automatically tune the value-of-information exploration rate so as to guarantee convergence to an optimal policy in the limit.  Informed by early experimental studies, we bounded the exploration rate between 0.01 and 5.01 and used a path-following policy accuracy of 0.001.

\subsection*{\small{\sf{\textbf{6.2$\;\;\;$Reinforcement Learning Results and Discussions}}}}

{\small{\sf{\textbf{Motion-Primitive Learning Performance.}}}} In what follows, we quantify the agent performance and discuss the underlying agent behaviors that emerge when planning with annotated motion primitives.  We then cover how the agents' ability to solve the two tasks changes when non-labeled primitives are employed.  Both of these cases correspond to the situation where partial prior task knowledge is encoded by the motion primitives.  The agents have the option of generalizing beyond these partial solutions, though, by interweaving arbitrary kinematics configurations.

Our simulation results for the two object-manipulation tasks are presented in figures 6.1 and 6.2.  An example of an action sequence that successfully completes the crate-pushing task is provided in figure 6.1(a).  As shown in figure 6.1(a), once the agent is both on the proper side of the crate and close enough to it, he can bend over to gain sufficient leverage.  This behavior is implemented by two labeled motion primitives.  In the remainder of the sequence, the agent relies on five labeled motion primitives to push the crate toward the goal.  These motion primitives correspond to hunched-over, straight-line walking actions.  Additional motion primitives were used to turn and walk toward the crate; these actions have been cut from the sequence shown in figure 6.1(a).  The agent also occasionally utilized non-primitive kinematics configurations to turn and orient itself properly in the direction of the crate.  An instance of a successful action sequence for the second task is shown in figure 6.2(a).  For this task, the agent initially relied on around fifteen primitives devoted to circular wiping motions to clear the small-scale, easy-to-clean regions.  It shifted to using both horizontal and vertical wiping primitives when the larger, harder-to-clean sections began to appear.  Ambulatory primitives were also chosen to maneuver to and around the wall.

\setcounter{figure}{0}
\begin{figure*}[t!]
\centering \vspace{-0.25cm} \hspace{-0.05cm} \includegraphics[width=6.5in]{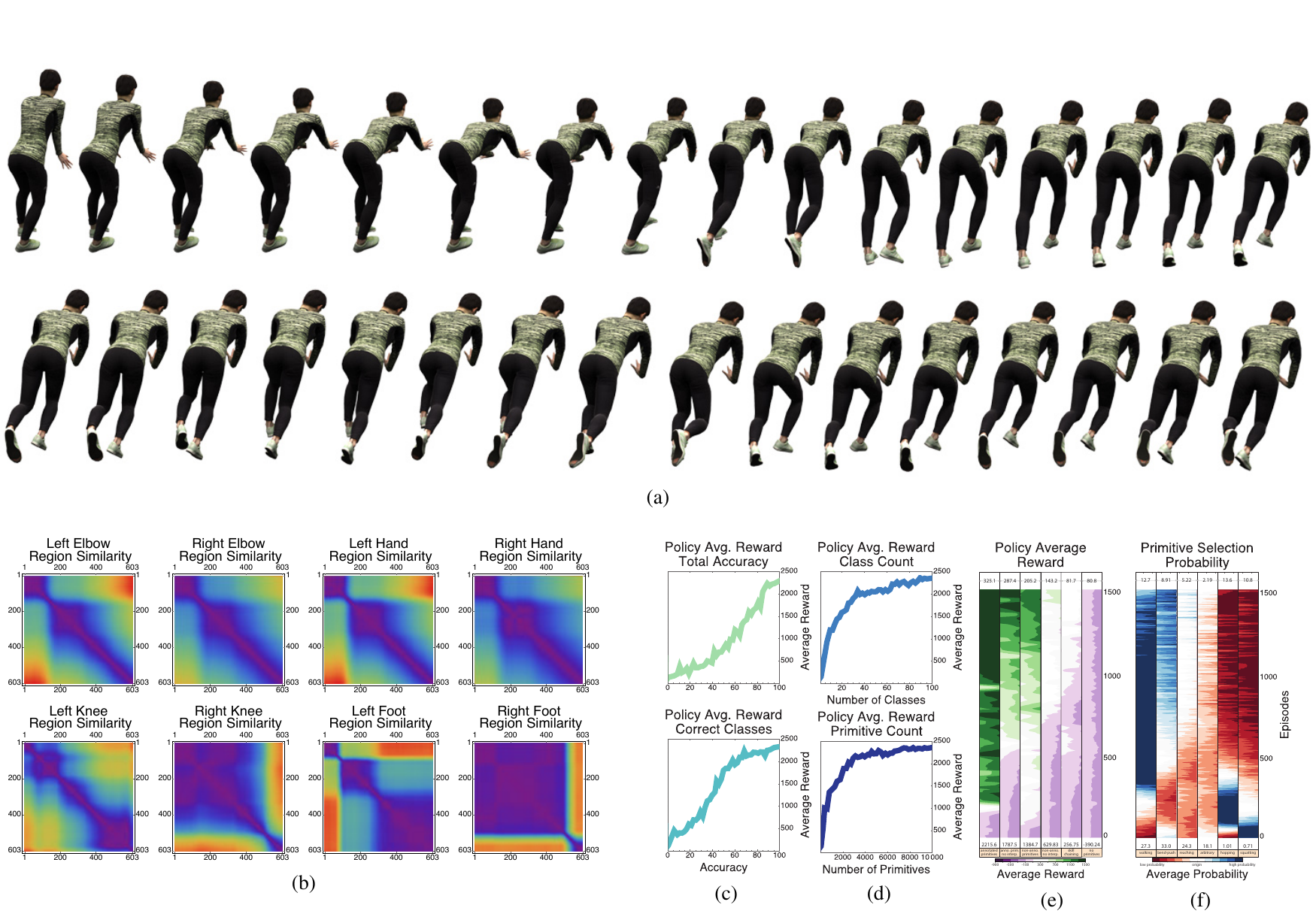}\vspace{-0.125cm}
\caption{Reinforcement learning results for the box-pushing object manipulation task.  (a) Subsequence of the motion trajectory produced by the best-reward policy.  In this trajectory, the actor partly bends over to gain sufficient leverage to slide the heavy box toward the goal location.  This trajectory resulted from a combination of seven motion primitives from four action classes.  Note that the box is not shown, since we focus just on the underlying motions.  (b) Plots of the pose-pose similarity over time for the subsequence shown in (a).  The color scheme for these plots is the same as in the previous figures.  These similarity plots show that the transition from one motion primitive to the next is smooth.  There are no visible discontinuities in the kinematics configurations over time.  (c) Plots of the total annotation accuracy and the per-class accuracy on the expected policy performance.  These plots indicate that the number of nearly-correctly annotated classes is a better predictor of performance than simply the overall annotation accuracy.  (d) Plots of the policy performance in terms of the number of unique action classes and the overall number of primitives per class, assuming that all of the primitives are accurately annotated.  Past a certain threshold, increasing the number of classes does little to improve performance; likewise, including redundant primitives in each class does little to enhance the obtainable policy rewards.  (e) Horizon plot of the average policy reward over the number of episodes when using annotated motion primitives both with and without experience interpolation, non-annotated primitives both with and without experience interpolation, and no primitives.  Purple colors indicate poor policy performance, with darker shades indicating negative reward scores.  Green colors denote high policy performance, with darker shades denoting higher rewards.  (f) Horizon plot of the normalized probabilities for choosing a specific class of annotated primitives as a function of the number of episodes.  Blue colors indicate a high probability of choosing a primitive class, while red colors suggest a lower chance of choosing that primitive class at a given episode.  Note that the chance of choosing a primitive class is averaged with respect to both the runs in a given episode and the entries of the policy at that episode.  The task-pertinent primitive probabilities thus typically lag behind the improvement in average rewards.\vspace{-0.35cm}}
\label{fig:fig6.1}
\end{figure*}

Although the action sequences needed to complete these two tasks appear simple, a great many episodes were needed to uncover the right selection and order of primitives to consistently create them.  As illustrated in figure 6.1(f) and 6.2(f), the policies initially emphasized the use of primitives from the turning and walking classes, as simply being within proximity of the objects would yield positive rewards.  Other primitive types, such as hopping, were briefly considered for the first task, as they would also enable the agent to get close to the objects; the time taken to do so was much longer, however, which discouraged their long-term use.  The policies later switched to executing more manipulation-based primitives once the agents had acquired enough knowledge about both the environment dynamics and the reward structure to better understand how to quickly complete the tasks.  Non-primitive kinematics configurations also became increasingly important at the later stages of training.  The shift in importance between the different primitive classes coincided with marked improvements in the overall rewards, as highlighted in figures 6.1(e) and 6.2(e).

When using non-annotated primitives, the implemented agent behaviors resembled those when using annotated primitives.  For the first task, the agent learned to head to the crate, bend over, and slowly push it toward the goal.  However, interspersed between these three dominant actions were either periods of inactivity or spurious movements, both of which prevented the agent from completing the task within the allotted time.  The reward plots in figure 6.1(e) corroborate this claim.  For the second task, the agent also determined that moving toward the wall and wiping it was an effective strategy.  Unlike in the annotated case, though, the agent defaulted to continuously wiping the entire wall with circular motions, regardless of where the grime regions were located.  This wasted significant amounts of time and often led to the agent not receiving the time-sensitive bonus reward.  The resulting performance was therefore worse than when the actions were composed of annotated primitives, as shown in figure 6.2(e).

The resulting policy quality was influenced greatly by two factors.  The first was the primitive annotation rate.  As shown in figures 6.1(d) and 6.2(d), the overall primitive annotation accuracy was crucial for discerning task-pertinent primitives.  The total number of correctly annotated classes was, in comparison, a better predictor of policy performance.  This is because if the primitives in each class were properly annotated, the agent could completely ignore them if they deemed unsuitable for a particular task.  By introducing annotation errors, the agent may have to exhaustively try each primitive.  When this occurs, the learning rate, and hence the corresponding accrued rewards, is subpar to that when non-annotated primitives are used as the action choices; this is highlighted in figure 6.1(e) and 6.2(e).  The second influential factor was the number of action classes and the number of primitives per class.  In figure 6.1(d) and 6.2(d), we show the effects of randomly excluding a certain number of action classes.  It can be seen that small amount of action classes increases the early policy performance, since the agent can more quickly explore the choices.  Too few classes is detrimental, though, since there may not be appropriate partial solutions upon which the agent can draw.  Likewise, reducing the number of primitives per class has a beneficial impact, up to some application-dependent threshold.  Exceeding that threshold significantly decreases policy performance.

\begin{figure*}[t!]
\centering \vspace{-0.25cm} \hspace{-0.05cm} \includegraphics[width=6.5in]{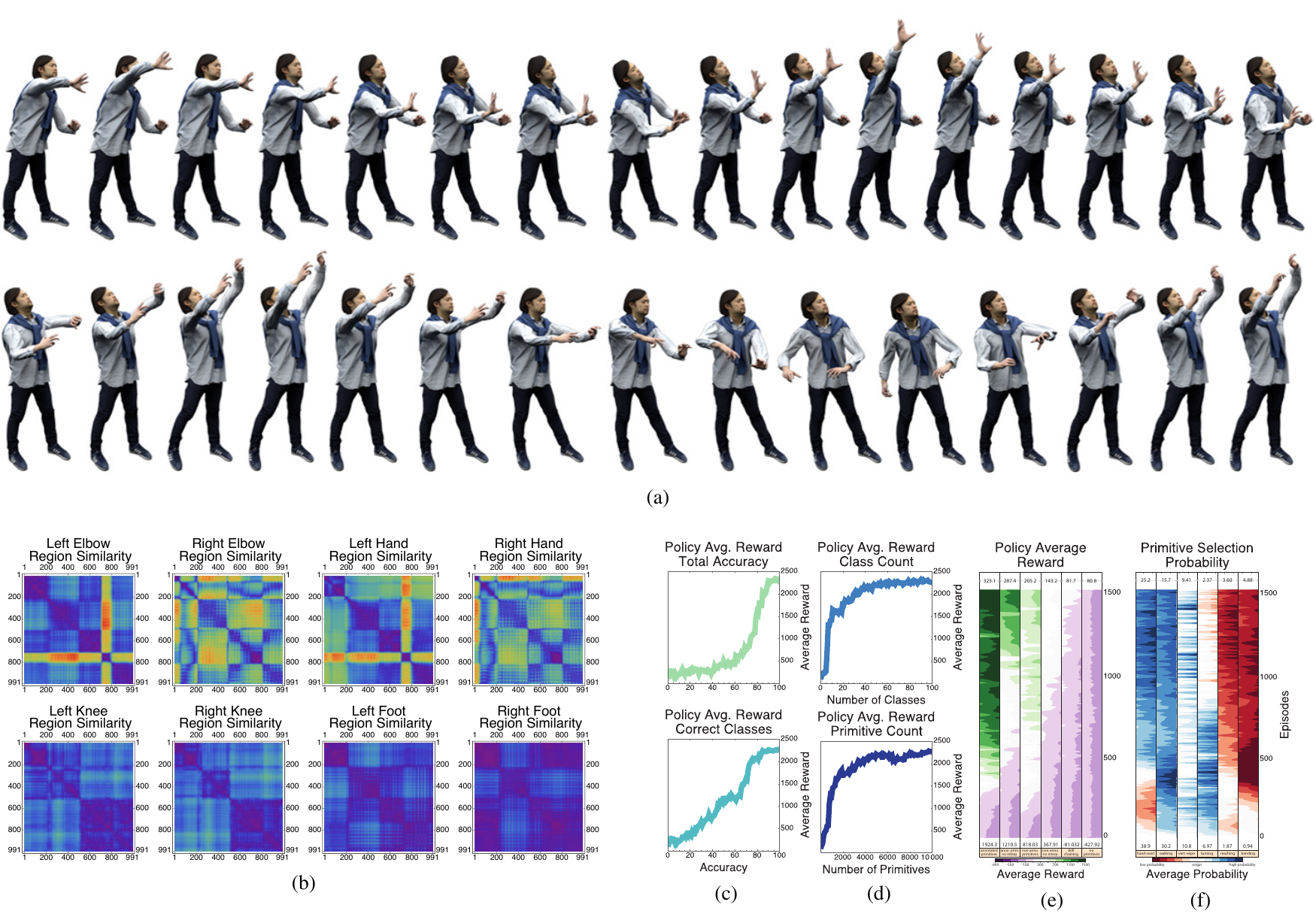}\vspace{-0.15cm}
\caption*{Reinforcement learning results for the wall wiping object manipulation task.  (a) Subsequence of the motion trajectory produced by the best-reward policy.  In this trajectory, the actor remains mostly in place and performs a series of circular wiping motions during the first half of the sequence.  In the second half, the actor wipes vertically.  This trajectory resulted from a combination of twenty-five motion primitives from four action classes.  Note that neither the wall nor the rag are shown, since we focus just on the underlying motions.  For descriptions of the plots in (b)--(f), see figure 6.1.\vspace{-0.35cm}}
\label{fig:fig6.2}
\end{figure*}

\vspace{0.15cm}{\small{\sf{\textbf{Comparative Learning Performance.}}}} The above findings indicate that motion primitives are effective for producing high-quality policies, since they encode partial task solutions.  To provide context for these results, we compare against standard semi-Markov $Q$-learning, where the actions are taken to be individual kinematics configurations, not sequences of configurations.  This corresponds to the general learning case where the agent has no prior knowledge that it can exploit and must discern how to solve a task solely through its interactions with the environment.  We also consider skill discovery and chaining \cite{KonidarisG-coll2009a,KonidarisG-coll2010a}, which fragments short-term kinematics sequences that arise during learning and strings them together.

Results for standard semi-Markov $Q$-learning are presented in figures 6.1(e) and 6.2(e).  The policies implemented only rudimentary and partial solutions for both tasks.  For the first task, the agent did not learn to ever push the heavy crate before the episode threshold was reached.  While it eventually learned to sometimes move toward the crate, it did so in a haphazard fashion; random strings of kinematics configurations were often interspersed with brief periods of locomotive behaviors.  Doubling, tripling, and even quadrupling the episode threshold did little to improve the task-solving capability of the agent, which likely stems from the incredibly high action-space dimensionality.  We witnessed parallel outcomes for the second task.  That is, the agent would sometimes randomly move toward and face the wall.  It did not, however, learn that there is a relationship between wiping-like arm and hand movements and rewards.  Including substantially more episodes did little to improve the agent's understanding of and ability to complete this task.  These findings are partly corroborated by the reward plots in figures 6.1(e) and 6.2(e), which highlight that, on average, the agent accrued little to no positive rewards for each episode.

Skill discovery and chaining faired somewhat better than semi-Markov $Q$-learning, as shown in figures 6.1(e) and 6.2(e).  Unlike the former case, the agents learned to more reliably move toward the crate, for the first task, and head toward the wall, for the second task.  This was due to the identification and utilization of basic locomotive skills, such as taking steps, which yielded moderate reward increases over semi-Markov $Q$-learning later during training.  The agents still did not learn how to finish much of the tasks, though, within the prescribed number of episodes.  While they sometimes learned to head to and push up against the crate, they did not do so consistently enough to move it much toward the goal location.  Likewise, the agents did not discover that the sponge could be used to wipe the wall.  These shortcomings were a byproduct of multiple factors.  First, the change-point temporal segmentation approach used by Konidaris et al. \cite{KonidarisG-coll2009a,KonidarisG-coll2010a} could not scale well to the highly articulated humanoid models; it often produced sub-par segmentations.  The incredibly high action-space dimensionality also led to difficulties: the agents could not adequately sample the action choices and experience a sufficient number of diverse state transitions within the limited number of episodes.  Doubling, tripling, and quadrupling the episode count led to marginal improvements in the average rewards, but they were nowhere near those obtained when using pre-specified motion primitives.

Both of these approaches were also hampered by the lack of constraints for choosing successive actions.  That is, the agents have the possibility of choosing an arbitrary kinematics configuration that can vastly differ from the current one.  Without any action regularization, it can be arduous to learn even simple behaviors, such as locomotion.

\vspace{0.15cm}{\small{\sf{\textbf{Reinforcement Learning Discussions.}}}} Through these experiments, we have quantitatively demonstrated that planning with motion primitives is advantageous.  With a sufficiently large, annotated library, semi-Markov $Q$-learning with motion primitives was shown to significantly outperform conventional $Q$-learning in far fewer episodes.  As well, if the primitives are accurately annotated, then better-performing policies could be found, with fewer training episodes, than when no labels were considered.

There are a few reasons why semi-Markov $Q$-learning with motion primitives can form high-performing policies more quickly than conventional $Q$-learning for the chosen tasks.  Foremost, the motion primitives that we consider encode task-specific knowledge and hence sidestep the need to recreate analogous movements during learning.  Standard semi-Markov-based $Q$-learning, in contrast, has no such prior knowledge that it can leverage; it must formulate task-pertinent kinematics sequences solely through its limited experiences with the environment.  While this style of learning without prior knowledge can be beneficial, it is not for those problems where abundant task information can be extracted prior to the learning process and suitably altered and utilized during it.  It is also not overly conducive to rapidly uncovering policies, since there are not any action-selection constraints; as well, there is no guidance on how to choose successive actions.  Motion primitives naturally impose such constraints, as they dictate a sequence of actions to be taken across a state series.  This simplifies exploration due to a reduction of the action space.

An advantage of using annotated motion primitives is that they achieve additional learning speed-ups compared to the non-annotated case.  This occurs because the agent exploits label information during the hierarchical action-selection process, allowing it to ignore entire classes of primitives if it finds that subsets of them do not aid in completing a task.  Such functionality further simplifies the action-search process.  As an example, in the wall wiping scenario, the agent discovered that some of the ambulatory primitives did not improve the obtainable long-term rewards.  It eschewed using any of those primitives, after trying a few of them from each action class, and quickly switched to those that contained upper body and arm movements.  In the non-annotated primitive case, the agent had to consider each motion primitive in the locomotive class multiple times before arriving at a similar conclusion.  This was because the agent had to understand the effect of each primitive on the obtainable rewards, since it lacked prior knowledge about the role and content of those primitives.

As we noted above, our differential-geometric annotation scheme achieved near-perfect recognition rates.  This was crucial for realizing the learning speed-ups that were observed in our experiments.  Alternate annotation techniques did poorly, which negated most of the learning-rate improvements compared to when non-annotated motion primitives were employed.  In some instances, the policy performance was lower than in the non-annotated case at various stages during training.  Upon further investigation, we found that the speed-ups diminished due not to the overall number of misclassified primitives, but rather the ratio of correctly to incorrectly recognized primitives in task-relevant action classes.  If too many primitives for a particular action were misclassified, then the agent often temporarily ignored that class, even if it was useful for completing a task.  The agent would then waste many episodes exploring alternate, non-useful action classes before reverting to systematically trying each of the primitives.  By that time, agents relying on non-annotated primitives often had more thoroughly surveyed all of the action choices and thus could better discern how to complete the object-manipulation tasks.  

Our motion-primitive-based approach relies on extracting and labeling the macro-actions, in an offline fashion, prior to their use for learning.  This is not the only way in which primitives could be defined, though.  An alternative would be to perform a process similar to what is used in skill chaining: identifying and segmenting motion trajectories as the agent interacts with the environment and understands the task reward structure.  Such an online scheme has the advantage that the resulting primitives would likely be applicable for completing related objectives.  A downside would be that the agent needs to solve the time-consuming and challenging problem of discerning what kinematics configuration works well for each state in a sequence without any prior knowledge.  We have partly sidestepped this issue with our approach, as task-specific domain knowledge is typically available in the primitives across multiple state-action pairs.  Another option would entail suitably altering portions of pre-specified primitives, during learning, so that they are more appropriate for a given task.  This has the benefit of still exploiting prior domain knowledge wherever possible with the flexibility of adapting to objectives that diverge from those in which the primitives were derived.  We will investigate this latter option in our future endeavors.

Just as accurately annotating motion primitives is important for performance, so too is having a diverse library of macro-actions, particularly if they either are not modified or are only changed slightly during learning.  In fact, for many application domains, it is better to have an abundance of disparate primitives per class than it is to have too few.  Utilizing more primitives naturally leads to solving an increasingly complicated search problem, since the agent has more action choices.  It does, however, permit better approximating the action-value function and will, with a sufficiently large number of primitives, allow the value-function estimates to approach those for the case where the kinematics configurations can be arbitrary.  This is because the agent has multiple ways in which it can potentially complete the assigned tasks.

\subsection*{\small{\sf{\textbf{7$\;\;\;$Conclusions}}}}\addtocounter{section}{1}

In this paper, we have proposed an approach for example-based annotation of movements and actions.  It is based on a descriptor that compares the changes in entities' three-dimensional poses and locations over time.  From this descriptor, we obtain a self-similarity matrix, which characterizes how similar a pose is at a given time step to all other instances in a single motion sequence.  We have found that this matrix encodes a great deal of information about the underlying movements.  As well, the similarities remain unchanged when the pose kinematics are translated, rotated, and scaled.  Invariance to affine and perspective transformations is possible with some modifications.

The descriptor that we developed is valid for multiple dimensions.  Spatial and temporal correspondences could also be resolved.  We exploited these properties to compare pairs of motion sequences, which we considered to be four-dimensional, space-time shapes.  In such situations, the descriptor returns a single value, versus a matrix.  We interpret this single value to be a measure of how related one movement sequence is to another sequence.  This allowed us to annotate the motion primitives using similarity-based classification schemes like weighted nearest-neighbor classifiers.

In our motion primitive annotation experiments, we determined that our approach is able to obtain a high annotation accuracy for human movements.  Our approach is mostly stable across camera viewpoint changes and can tolerate some deviations in the movements of sequence pairs.  It is also relatively insensitive to the motion speed.  Noise caused by the pose estimation process is also mostly ignored; it only impacts the annotation rate if it significantly deforms the pose throughout the entire action sequence.  It can perform well when only either a few labeled action examples are used or just a single labeled example is provided.  The annotation rate is on par with state-of-the-art deep neural networks that have access to much larger amounts of data.  This indicates that a great deal of intrinsic information about the motion primitives is being efficiently extracted and encoded by our approach.

Our motivation for annotating primitives was to decrease the time needed to acquire various skills during semi-Markov-decision-process-based reinforcement learning.  In our simulations, we showed that learning with annotated primitives was more effective versus non-labeled ones.  We attributed this improvement to the fact that using the label in the hierarchical action-selection process enables the learning process to quickly filter out entire classes of primitives that are not useful for completing a task.  The improvement amount depends greatly on the annotation accuracy, though.  Our differential-geometric approach provided a near-perfect action and motion recognition rate on the underlying human movement datasets, which greatly improved the policy quality compared to when non-annotated primitives were used.  When the annotation rate was artificially lowered, we found that the improvement gap naturally decreased.  The greatest predictor of policy performance was not simply the percentage of correct labels, but rather the percentage of correctly labeled primitives in classes that were useful for completing tasks.

We additionally demonstrated that using motion primitives can be more effective than attempting to empirically determine a potentially unique action for each state.  Good policies are often uncovered much earlier during learning in the former case.  There are a few reasons for this.  Foremost, using motion primitives constrains the search process in the sense that they reduce the number of possible action choices compared to the situation where arbitrary kinematics sequences can be specified.  Additionally, motion primitives bootstrap the learning process.  Provided that the motion primitives are relevant for the chosen domain, they encode possible partial solutions for various tasks.  This preempts having to re-discover such motion sequences purely through the agent's interactions with the environment, which saves time.  The agent can therefore focus on the simpler problem of discerning what motion templates work well for certain situations.

In our future work, we will focus on discerning how to best interrupt in-progress motion primitives and altering them to react to environment changes.  Moreover, we will show how the learning problem can be modified to intelligently alter and aggregate parts of multiple motion primitives into a single, new primitive.  Both efforts should reduce the number of primitives needed to complete complex tasks using reinforcement learning.

\renewcommand*{\bibfont}{\raggedright}
\renewcommand\bibsection{\subsection*{\small{\sf{\textbf{References}}}}}
{\singlespacing\fontsize{9.75}{10}\selectfont \bibliography{sledgebib}
\bibliographystyle{IEEEtran}}

\end{document}